\pdfoutput=1

\documentclass[11pt]{article}

\usepackage[]{ACL2023}

\usepackage{times}
\usepackage{latexsym}

\usepackage[T1]{fontenc}

\usepackage[utf8]{inputenc}

\usepackage{microtype}

\usepackage{times}
\usepackage{latexsym}

\usepackage{comment}
\usepackage{rotating}
\usepackage[utf8]{inputenc}
\usepackage{times}
\usepackage{latexsym}

\usepackage{float}
\usepackage{graphicx}
\usepackage{subfigure}
\usepackage{caption}
\usepackage{amsmath,amssymb}
\usepackage{mathtools}
\usepackage{booktabs}
\usepackage{multirow}
\usepackage{multicol}
\usepackage{siunitx}
\usepackage{adjustbox}
\usepackage{pifont}

\usepackage{enumitem}
\usepackage{color}

\usepackage[T1]{fontenc}

\usepackage[utf8]{inputenc}

\usepackage{microtype}

\newcommand*{\affaddr}[1]{#1} 
\newcommand*{\affmark}[1][*]
{\textsuperscript{#1}}

\newcommand{\specialcell}[2][c]{%
  \begin{tabular}[#1]{@{}c@{}}#2\end{tabular}}

\title{Exploring the Potential of ChatGPT on Sentence Level Relations: A Focus on Temporal, Causal, and Discourse Relations}

\author{
Chunkit Chan\affmark[1],
Cheng Jiayang\affmark[1],
Weiqi Wang\affmark[1],
Yuxin Jiang\affmark[2],\\
\textbf{Tianqing Fang\affmark[1],
Xin Liu\affmark[1],
Yangqiu Song\affmark[1]}\\
\affaddr{\affmark[1]Department of Computer Science and Engineering, HKUST, Hong Kong SAR, China}\\
\affaddr{\affmark[2]Information Hub, HKUST (GZ), Guangzhou, China} \\
\texttt{\{ckchancc, yqsong\}@cse.ust.hk} \\
}

\begin{document}
\maketitle
\begin{abstract}
This paper aims to quantitatively evaluate the performance of ChatGPT, an interactive large language model, on inter-sentential relations such as temporal relations, causal relations, and discourse relations. 
Given ChatGPT's promising performance across various tasks, we proceed to carry out thorough evaluations on the whole test sets of 11 datasets, including temporal and causal relations, PDTB2.0-based, and dialogue-based discourse relations. To ensure the reliability of our findings, we employ three tailored prompt templates for each task, including the zero-shot prompt template, zero-shot prompt engineering (\textbf{PE}) template, and in-context learning (\textbf{ICL}) prompt template, to establish the initial baseline scores for all popular sentence-pair relation classification tasks for the first time.\footnote{The code and prompt template are available at~\url{https://github.com/HKUST-KnowComp/ChatGPT-Inter-Sentential-Relations}.} Through our study, we discover that ChatGPT exhibits exceptional proficiency in detecting and reasoning about causal relations, albeit it may not possess the same level of expertise in identifying the temporal order between two events. While it is capable of identifying the majority of discourse relations with existing explicit discourse connectives, the implicit discourse relation remains a formidable challenge. Concurrently, ChatGPT demonstrates subpar performance in the dialogue discourse parsing task that requires structural understanding in a dialogue before being aware of the discourse relation.
\end{abstract}

\section{Introduction}

With the proliferation of computational resources and the availability of extensive text corpora, the expeditious advancement of large language models (e.g., ChatGPT~\cite{openai2022chatgpt} and GPT-4~\cite{DBLP:journals/corr/abs-2303-08774}) have prominently showcased their emergence ability resulting from the scaling up model size. Techniques such as instruction tuning~\cite{wei2022finetuned} and reinforcement learning from human feedback~\cite{ouyang2022training} have further fortified LLM with sophisticated language understanding and logical reasoning proficiencies. Therefore, these large language models (LLMs) demonstrate remarkable few-shot, even zero-shot learning abilities in performing various tasks. Recent studies have extensively and comprehensively evaluated ChatGPT's performance on numerous language understanding and reasoning tasks, revealing that its superior performance in zero-shot scenarios when compared to other models~\cite{
DBLP:journals/corr/abs-2303-12712,
DBLP:journals/corr/abs-2302-04023,
DBLP:journals/corr/abs-2301-08745, 
DBLP:journals/corr/abs-2302-10724}. Besides, ChatGPT has also shown impressive powers in data annotations and has proven to be more cost-efficient than crowd-workers for several annotation tasks~\cite{DBLP:journals/corr/abs-2304-06588, DBLP:journals/corr/abs-2303-15056}.
Whilst the success of ChatGPT has been witnessed, certain obstacles persist unaddressed. Previous research has discussed the associated ethical implications and privacy concerns~\cite{DBLP:journals/corr/abs-2212-09292, DBLP:journals/corr/abs-2302-00539,DBLP:journals/corr/abs-2310-10383,DBLP:journals/corr/abs-2311-04044}. Moreover, ChatGPT's shortcomings include but are not limited to the lack of planning~\cite{DBLP:journals/corr/abs-2303-12712}, the inability to perform complex mathematical reasoning~\cite{DBLP:journals/corr/abs-2301-13867}, and fact validation~\cite{DBLP:journals/corr/abs-2302-13817, 
wang2023survey,
DBLP:journals/corr/abs-2302-04023}. Consequently, it is still under discussion whether large language models possess the capacity to comprehend text beyond surface forms as humans.

To comprehend the natural language text at a deeper level, it is crucial for an LLM to capture and understand the higher-level inter-sentential relations from the text, which involves mastering more complex and abstract relations beyond surface forms. These inter-sentential relations, such as temporal, causal, and discourse relations between two sentences, are widely used to form knowledge that has been proven to benefit many downstream tasks~\cite{
DBLP:conf/emnlp/DaiH19,
DBLP:conf/acl/TangLL0H0XX20,
DBLP:journals/corr/abs-2302-09715,
DBLP:journals/corr/abs-2304-03754}.
In this study, we quantitatively evaluate the performance of ChatGPT in tasks that require an understanding of sentence-level relations, including temporal relation (Section \ref{sec:temporal}), causal relation (Section \ref{sec:causal}), and discourse relation (Section \ref{sec:discourse}). Under three standard prompt settings\footnote{Zero-shot prompting (denoted by \textbf{Prompt}), zero-shot prompt engineering (\textbf{PE}), and in-context learning (\textbf{ICL}). Prompt examples are shown in Appendix \ref{sec:appendix_prompts}.},
we conduct extensive evaluations on the \textit{whole} test sets of 11 datasets regarding these relations.\footnote{We exclude entailment or NLI tasks because they have already been evaluated in previous studies~\cite{DBLP:journals/corr/abs-2302-10724, DBLP:journals/corr/abs-2302-10198}.} 
Furthermore, we conducted an in-depth study on the various intra-relations of each inter-sentential relation (e.g., \textit{Before} and \textit{After} relation in Temporal relations) and assessed the performance of the ChatGPT on these specific intra-relations. The detailed relation-wise performance is shown in Figure~\ref{fig: relation_wise_comparison_ALL}.
The primary insights drawn from the analysis of quantitative assessments are as follows\footnote{All evaluations were performed in April 2023 using the OpenAI API (\textit{gpt-3.5-turbo-0301 model}), and similar performance was observed in the latest model (\textit{"gpt-3.5-turbo-1106"}).}:

\begin{itemize}
    \item \textbf{Temporal relations:} ChatGPT \textbf{has difficulty} in identifying the temporal order between two events, which could be attributed to inadequate human feedback on this feature during the model’s training process.
    \item \textbf{Causal relations:} ChatGPT \textbf{exhibits strong} performance in detecting and reasoning about causal relationships, particularly on the COPA dataset. It also outperforms fine-tuned RoBERTa on two out of three benchmarks.
    \item \textbf{Discourse relations:} Explicit discourse relations can \textbf{be easily recognized} by ChatGPT thanks to the explicit discourse connectives in context. However, it \textbf{struggles} with the absence of connectives for implicit discourse tasks, particularly with the link and relation prediction in dialogue discourse parsing.
\end{itemize}
We aspire to contribute to the research community through our evaluations and discoveries. By sharing the result, we intend to offer valuable insights to others in the relevant fields.

\section{Related Work}

\paragraph{Large Language Model}
With the increase of computational resources and available text corpora, the research community has discovered that large language models (LLMs) show an impressive ability in few-shot, even zero-shot learning with scaling up~\cite{brown2020language,kaplan2020scaling,wei2022finetuned, DBLP:journals/corr/abs-2305-12870}.
Besides, instruction tuning~\cite{wei2022finetuned} and reinforcement learning from human feedback~\cite{ouyang2022training} also empower LLM with complicated language understanding and reasoning.
Recently, ChatGPT~\cite{openai2022chatgpt} and GPT-4~\cite{DBLP:journals/corr/abs-2303-08774} have achieved remarkable performance on a wide range of natural language processing benchmarks, including language modeling, machine translation, question answering, text completion, commonsense reasoning, and even human professional and academic exams.
These achievements have garnered significant attention from academia and industry, and many efforts have been made to estimate the potential of artificial general intelligence (AGI)~\cite{DBLP:journals/corr/abs-2302-04023, DBLP:journals/corr/abs-2304-06364,DBLP:journals/corr/abs-2301-13867,DBLP:journals/corr/abs-2302-04752,DBLP:journals/corr/abs-2304-05454,wang2024candle}.
It is crucial for the research community to continue exploring the capabilities of LLMs in various directions and tasks for further development of NLP.

\paragraph{Temporal Relation}
Temporal relation extraction aims to detect the temporal relation between two event triggers in the given document \cite{pustejovsky2003timeml}.
It is crucial for many downstream NLP tasks since reasoning over temporal relations plays an essential role in identifying the timing of events, estimating the duration of activities, and summarizing the chronological order of a series of occurrences \cite{ning2018matres}. 
There exists a recent work that evaluates ChatGPT's ability on zero-shot temporal relation extraction \cite{DBLP:journals/corr/abs-2304-05454}. However, their manually designed prompts acquire unsatisfiable performance, and the capability of ChatGPT equipped with in-context learning has not been explored. Therefore, this work also includes the temporal relation tasks, and our results can complement and validate each other with \citet{DBLP:journals/corr/abs-2304-05454}.

\paragraph{Causal Relation} 
Causal reasoning involves the identification of causality, which refers to the connection between a cause and its corresponding effect~\cite{DBLP:conf/ijcai/Bochman03}.
NLP models that can reason causally have the potential to improve their ability to understand language, as well as to solve complex problems in various fields, such as physical reasoning~\cite{DBLP:conf/acl/AtesAYKKEEGY22}, event extraction~\cite{DBLP:conf/coling/CuiSCLLS22}, question-answering~\cite{DBLP:conf/naacl/ZhangZL022, DBLP:conf/emnlp/SharpSJCH16}, and text classification~\cite{DBLP:conf/aaai/ChoiJHH22}.
Although \citet{DBLP:journals/corr/abs-2301-13819} has analyzed ChatGPT's performance in a medical causality benchmark, no prior research has conducted a comprehensive study on the ability of large language models to reason upon causal relations.

\paragraph{Discourse Relation}
Discourse relation recognition is a vital task in discourse parsing, identifying the relations between two arguments (i.e., sentences or clauses) in the discourse structure. 
It is essential for textual coherence and is regarded as a critical step in constructing a knowledge graph~\cite{DBLP:conf/www/ZhangLPSL20, DBLP:journals/ai/ZhangLPKOFS22} and various downstream tasks involving more context, such as text generation~\cite{DBLP:conf/naacl/BosselutCHGHC18}, text categorization~\cite{DBLP:conf/acl/LiuOSJ20}, and question answering~\cite{DBLP:conf/acl/JansenSC14}. Explicit discourse relation recognition (EDRR) has already shown that utilizing explicit connective information can effectively determine the types of discourse relations~\cite{DBLP:conf/sigdial/VariaHC19}. In contrast, implicit discourse relation recognition (IDRR) remains challenging because of the absence of connectives. However, previous works have not systemically evaluated the ability of ChatGPT on these two discourse relation recognition tasks. Therefore, in this work, we assess the performance of this large language model (i.e., ChatGPT) on the PDTB-style discourse relation recognition task~\cite{DBLP:conf/lrec/PrasadDLMRJW08}, dialogue discourse parsing~\cite{asher2016discourse, li2020molweni}, and downstream applications on discourse understanding.

\section{Experimental Setting} 
We employ three customized prompt templates for each task: zero-shot setting, zero-shot with prompt engineering (\textbf{PE}), and the in-context learning (\textbf{ICL}) setting. The devised prompt template will serve as comprehensive and reliable baselines to exclude the variance of the prompt engineering and offer fair comparison baselines for all prevalent sentence-pair relation classification tasks.
The specific template details are presented in corresponding sections and Appendix~\ref{sec:appendix_prompts}. 

\begin{itemize}
    \item \textbf{ChatGPT$_{\text{Prompt}}$} refers to formulating the task as a multiple choice question answering problem and utilizing the prompt template in \citet{robinson2022leveraging} as a baseline.
    \item \textbf{ChatGPT$_{\text{Prompt Engineering}}$} refers to manually designing a more sophisticated prompt template based on the expert understanding of various tasks. 
    \item \textbf{ChatGPT$_{\text{In-Context Learning}}$} refers to the in-context learning prompting method inspired by \citet{brown2020language}. We manually select $C$ input-output exemplars from the train split and reformulate these examples into our prompt-engineered template, where $C$ is the number of classes. These well-selected examples for each category are distinguishable and easily understandable examples between each class. 
\end{itemize}


\section{Temporal Relation}
\label{sec:temporal}
Temporal relation extraction aims to determine the temporal order between two events in a text \cite{pustejovsky2003timeml}, which could be formulated as a multi-label classification problem.
In this section, we evaluate the temporal reasoning ability of ChatGPT on three commonly used benchmarks: TB-Dense \cite{cassidy2014tbdense}, MATRES \cite{ning2018matres}, and TDDMan \cite{naik2019tddiscourse} (details in Appendix~\ref{sec:experimental setting}).
To ensure compatibility with previous research, we employ the same data split and assess ChatGPT's performance on the entire test set.




\paragraph{Detailed Experimental Setting.}
In comparison to random guess, the supervised baseline BERT-base \cite{mathur2021timers}, and the supervised state-of-the-art model RSGT \cite{zhou2022rsgt}, we equip ChatGPT using three popular prompting strategies shown in Tables \ref{tab:temporal_TB-Dense_templete1}, \ref{tab:temporal_TB-Dense_templete2}, \ref{tab:temporal_MATRES_templete}, \ref{tab:temporal_TDDMan_templete1}, and \ref{tab:temporal_TDDMan_templete2} in Appendix~\ref{sec:appendix_prompts}. For ChatGPT$_{\text{Prompt Engineering}}$, we manually design a more sophisticated prompt template to remind ChatGPT to first pay attention to the temporal order as well as the two events, which largely boosts its prediction performance.


\begin{figure*}[!t]
\centering
\vspace{0.15in}
\includegraphics[width=\textwidth]{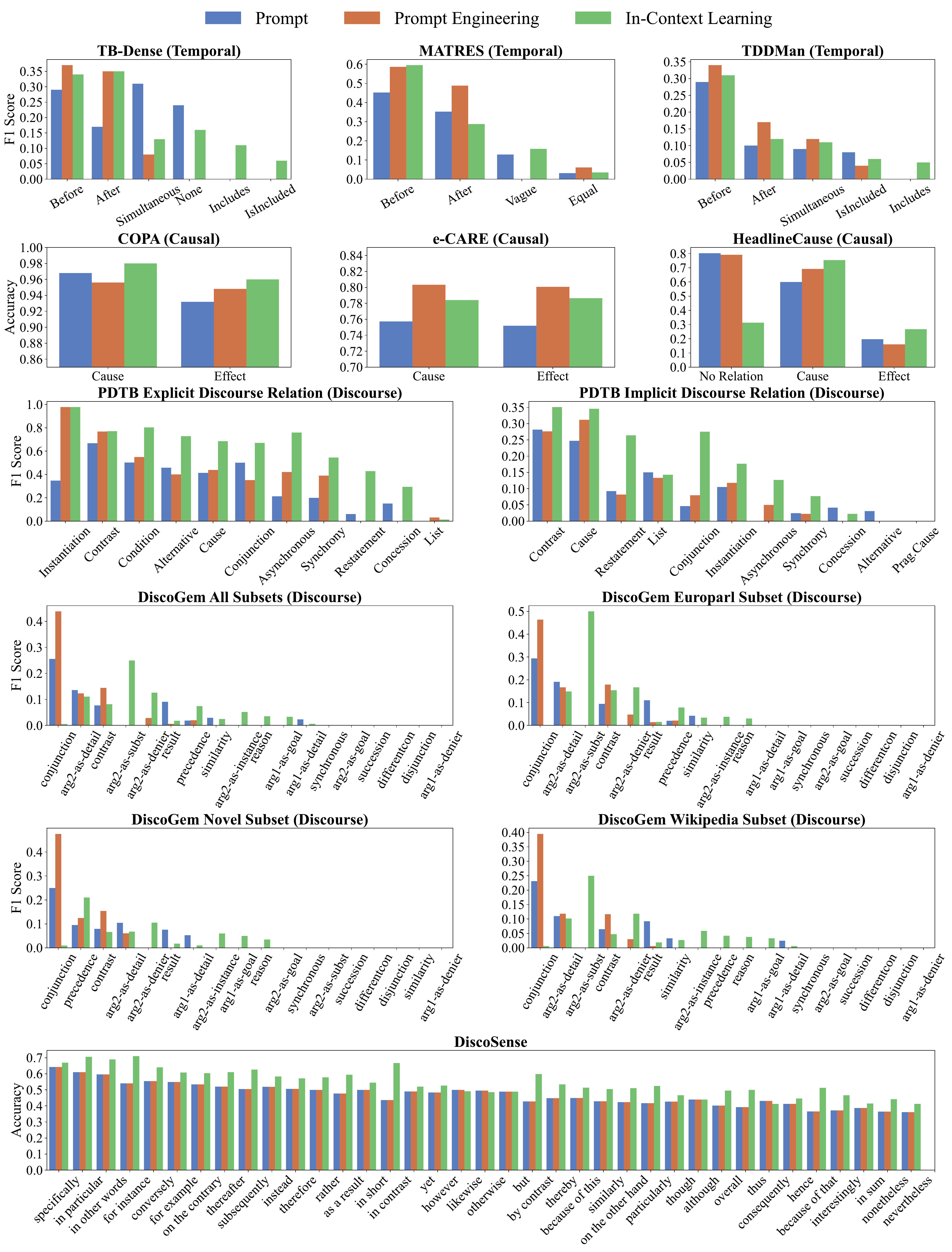}
\caption{Relation-wise performance comparison on temporal, causal, and discourse benchmarks by ChatGPT with different prompting methods. DiscoSense is a downstream task of discourse relations.}
\label{fig: relation_wise_comparison_ALL}
\vspace{0.8cm}
\end{figure*}

\begin{table}[!t]
\small
\centering
\setlength\tabcolsep{4pt}
\scalebox{0.85}{
\begin{tabular}{l|c|c|c}
\toprule
\multicolumn{1}{c|}{\textbf{Method}} & \textbf{TB-Dense} & \textbf{MATRES} & \textbf{TDDMan} \\ \midrule
Random      & 15.0    & 25.8   & 17.3                 \\
BERT-base   & 62.2    & 77.2   & 37.5                 \\
Fine-tuned SOTA & 68.7 & 84.0  & 45.5                \\ \midrule
ChatGPT$_{\text{Prompt}}$  & 23.3   & 35.0   & 14.1    \\
ChatGPT$_{\text{PE}}$ & 27.0  &  47.9  & 16.8 \\
ChatGPT$_{\text{ICL}}$ & 25.0  & 44.9 & 14.7    \\ \bottomrule
\end{tabular}
}
\vspace{-0.3cm}
\caption{
The Micor-F1 performance (\%) of ChatGPT on temporal relation extraction. 
}
\label{tab:temporal_experiment}
\end{table}

\paragraph{Experimental Result.} Table \ref{tab:temporal_experiment} presents the results of the experiment, where \textbf{ChatGPT lags behind fine-tuned models by more than 30\% on all three datasets}. This suggests that ChatGPT may not be proficient in identifying the temporal order between two events, which could be attributed to inadequate human feedback on this feature during the model's training process. 
Additionally, our advanced prompt engineering delivers superior performance compared to the standard prompting baseline, with an improvement of 3.7\%, 12.9\%, and 2.7\% on TB-Dense, MATRES, and TDDMan, respectively. Throughout our experiments, three significant observations emerged, which are worth noting:

(1) In temporal relation extraction tasks, ChatGPT's performance did not improve through in-context learning. The performance of in-context learning can be highly unstable across samples of examples, indicating that the process of language model acquiring information is idiosyncratic~\cite{DBLP:journals/corr/abs-2302-13539,DBLP:conf/emnlp/ZhangFT22}. A number of case studies are provided in Tables \ref{tab:temporal_TB-Dense_templete1}, \ref{tab:temporal_TB-Dense_templete2}, \ref{tab:temporal_MATRES_templete}, \ref{tab:temporal_TDDMan_templete1}, and \ref{tab:temporal_TDDMan_templete2} in Appendix~\ref{sec:appendix_prompts}. These tables display test examples formulated into three templates using the aforementioned prompting strategies and subsequently fed to ChatGPT for response generation. The results indicate that only prompt engineering yields correct answers. We explored the underlying reasons by examining label-wise F1 performance, as illustrated in Figure \ref{fig: relation_wise_comparison_ALL}. It appears that in-context learning enhances performance for more difficult-to-distinguish relations, such as \textit{INCLUDES} and \textit{IS\_INCLUDED}, but negatively impacts performance for more easily distinguishable relations, like \textit{BEFORE} and \textit{AFTER}.

(2) ChatGPT exhibits a tendency to predict the temporal relation between $event_1$ and $event_2$ as \textit{BEFORE}. This suggests a limited understanding of temporal order, given that the sequence of $event_1$ typically precedes $event_2$ within the text.

(3) In the context of long-dependency temporal relation extraction, ChatGPT is unsuccessful. As demonstrated in Table \ref{tab:temporal_experiment}, ChatGPT, when equipped with all three prompting strategies, performs worse than random guessing on TDDMan. This dataset primarily focuses on long-document and discourse-level temporal relations, with an example provided in Tables \ref{tab:temporal_TDDMan_templete1} and \ref{tab:temporal_TDDMan_templete2} in Appendix~\ref{sec:appendix_prompts}.


\section{Causal Relation} 
\label{sec:causal}
Causal reasoning is the process of understanding and explaining the cause-and-effect relationships between events~\cite{DBLP:conf/acl/CaoZ000CP20}. 
It involves identifying the factors that contribute to a particular result and understanding how changes in those factors can lead to different outcomes~\cite{DBLP:conf/acl/RothWNF18,DBLP:conf/emnlp/PontiGMLVK20}. 
In this paper, we assess the causal reasoning ability of LLMs by benchmarking their results on three existing causal reasoning datasets (COPA \cite{DBLP:conf/semeval/GordonKR12}, e-CARE \cite{DBLP:conf/acl/DuDX0022}, and HeadlineCause \cite{DBLP:conf/lrec/GusevT22}, details in Appendix~\ref{sec:experimental setting}) and quantitatively analyzing the results.
Our findings demonstrate that the LLM exhibits a robust ability to detect and reason about causal relationships, particularly those pertaining to cause and effect, without requiring advanced prompting techniques such as in-context learning.

\begin{table}[!t]
\centering
\small
\setlength\tabcolsep{4pt}
\scalebox{0.85}{\begin{tabular}{@{}l|c|c|c@{}}
\toprule
\multicolumn{1}{c|}{\textbf{Method}} & \textbf{COPA} & \textbf{e-CARE} & \textbf{HeadlineCause} \\
\midrule
Random & 50.0 & 50.0 & 20.0 \\
Fine-tuned RoBERTa & 90.6 & 70.7 & 73.5 \\
Fine-tuned SOTA & 100.0 & 74.6 & 83.5\\
 \midrule
ChatGPT$_{\text{Prompt}}$ & 94.8 & 74.8 & 71.4 \\
ChatGPT$_{\text{PE}}$ & 95.2 & 79.6 & 72.7 \\ 
ChatGPT$_{\text{ICL}}$ & 97.0 & 78.6 & 36.2 \\ 
\bottomrule
\end{tabular}
}
\caption{Experiment results (Accuracy \%) of fine-tuned RoBERTa and ChatGPT on causal reasoning benchmarks. 
}
\label{tab:causal_reasoning_result}
\vspace{-0.4cm}
\end{table}

\paragraph{Detailed Experimental Setting.}
For the baseline, we report the accuracy of \textit{random labeling} to reflect the character of each dataset and fine-tuned \textit{RoBERTa}~\cite{DBLP:journals/corr/abs-1907-11692} to show the power of fine-tuned pre-trained language models.
Accuracy is used as the evaluation metric to assess ChatGPT on three benchmarks using three different prompting techniques. 
The detailed prompts for three benchmarks are shown in Table~\ref{tab:copa_pemplate}, Table~\ref{tab:ecare_pemplate}, and Table~\ref{tab:headlinecause_pemplate} in Appendix~\ref{sec:appendix_prompts}, respectively. Table~\ref{tab:causal_reasoning_result} presents the results of our experiments. 
For the ChatGPT$_{\text{Prompt Engineering}}$, we use more sophisticated prompt designs that emphasize the explanation of the question setting (what is the relationship between the given event and its options) and the causal relations.

\paragraph{Experimental Results.} Notably, ChatGPT demonstrates exceptional performance on the COPA dataset and satisfactory performance on the other two datasets, outperforming fine-tuned RoBERTa on two out of three benchmarks and achieving comparable performance on the HeadlineCause dataset.
Our engineered prompt improves performance slightly across all benchmarks, while in-context learning enhances ChatGPT's ability to excel only on the COPA dataset but has a detrimental effect on the HeadlineCause dataset.
To gain deeper insights, we conduct relation-wise comparisons of ChatGPT's performance on all three benchmarks, specifically examining its accuracy in identifying \textit{cause} and \textit{effect} relationships under different prompting techniques.
The results are shown in Figure~\ref{fig: relation_wise_comparison_ALL}.
Using the engineered prompt and in-context learning prompt tends to yield the best performance on the COPA and e-CARE datasets. 
However, for the HeadlineCause dataset, while in-context learning improves ChatGPT's ability to identify \textit{cause} and \textit{effect} relationships, it also makes it harder for the model to discriminate \textit{no relation} entries.

In conclusion, our experiments demonstrate that \textbf{ChatGPT exhibits strong performance in detecting and reasoning about causal relationships, particularly those pertaining to cause and effect.}
Our results also indicate that using engineered prompts and in-context learning can enhance ChatGPT's performance across various benchmarks, sometimes surpassing supervised baselines. 
However, the effectiveness of these techniques varies depending on the dataset.
We hope this work can shed light on the strengths and limitations of ChatGPT in causal reasoning tasks and inform future research in this area.

\section{Discourse Relation}
\label{sec:discourse}
In this section, we evaluate ChatGPT on Discourse Relation recognition tasks, including \textit{PDTB-Style Discourse Relation Recognition}, \textit{Multi-genre Crowd-sourced Discourse Relation Recognition}, \textit{Dialogue Discourse Parsing}, and applications on discourse understanding. Apart from these datasets and tasks, we conduct the assessments of ChatGPT's performance on two downstream tasks which are shown in Appendix~\ref{sec:appendix_downstreamtask}.

\subsection{PDTB-Style Discourse Relation Recognition}

\paragraph{Detailed Experimental Setting.} Explicit discourse relation recognition aims to recognize the discourse relation between two arguments, with the explicit discourse markers or connectives (e.g., ``so'', and ``because'') in between. 
In comparison, the implicit setting identifies the discourse relation without connectives. 
The labels of these two tasks for each discourse relation in the PDTB2.0~\cite{DBLP:conf/lrec/PrasadDLMRJW08} follow the hierarchical classification scheme throughout the annotation process, annotated as a hierarchy structure (shown in Figure~\ref{fig:PDTB_Senses_Hierarchy} in Appendix). 
In this work, we evaluate ChatGPT's performance on PDTB 2.0 (Ji-setting~\cite{DBLP:journals/tacl/JiE15}), and the details are presented in Appendix~\ref{sec:experimental setting}.
The example of discourse relations in Figure~\ref{fig:Implict discourse relation example} in Appendix~\ref{sec:experimental setting} shows the \textit{Contingency} top-level class and \textit{Cause} second-level class. 
The details of three tailored prompt templates are provided in the Tables~\ref{tab:explicitdiscourse_templete_1},~\ref{tab:explicitdiscourse_templete_2},~\ref{tab:implicitdiscourse_templete_1}, and~\ref{tab:implicitdiscourse_templete_2} in Appendix~\ref{sec:appendix_prompts}.

For ChatGPT$_{\text{Prompt Engineering}}$, we manually designed a task-specified prompt as follows. 
Since the label of the PDTB2.0 dataset inherently forms the hierarchy, we utilized this label dependence to tailor a prompt template to predict the top-level class and second-level class simultaneously. 
Moreover, we select a representative connective for each discourse relation in the IDRR task, while the EDRR task already provides the explicit connectives for each instance.
Therefore, we use the label dependence and the selected connectives to guide the LLM to understand the sense of each discourse relation.

\begin{table}[!t]
\small
\centering
\setlength\tabcolsep{4pt}
\scalebox{0.85}{
\begin{tabular}{l|cc|cc}
\toprule
\multicolumn{1}{c|}{\multirow{2}{*}{\textbf{Method}}} & \multicolumn{2}{c|}{\textbf{Top}} & \multicolumn{2}{c}{\textbf{Second}} \\
& F1 & Acc & F1 & Acc \\
\midrule
Random & 25.12 & 25.70 & 7.30 & 9.19 \\
\citet{DBLP:conf/emnlp/ZhouLWCM22} & 93.59 & 94.78 & - & - \\
\citet{DBLP:conf/sigdial/VariaHC19} & 95.48 & 96.20 & - & - \\
\citet{DBLP:journals/corr/abs-2305-03973}& 95.64 & 96.73 &  - & - \\
\hline
ChatGPT$_{\text{Prompt}}$ & 34.94 & 39.38 & 31.92 & 43.26 \\
ChatGPT$_{\text{PE}}$  & 69.26 & 70.21 & 39.34 & 50.80 \\
ChatGPT$_{\text{ICL}}$ & 84.66 & 85.97 & 60.68 & 63.47 \\
\bottomrule
\end{tabular}
}
\caption{
The performance of ChatGPT performs on the explicit discourse relation recognition task of PDTB (\textit{Ji}) test set. 
}
\label{tab:ChatGPT_EXP_Performance}
\end{table}
\subsubsection{Explicit Discourse Relation Recognition}
\paragraph{Experimental Results.} 
In Table~\ref{tab:ChatGPT_EXP_Performance}, the performance shows that \textbf{ChatGPT can recognize each explicit discourse relation by utilizing the information from the explicit discourse connectives.} 
Furthermore, by utilizing the label dependence between the top-level label and the second-level label to design the prompt template, the performance of the top-level class increases significantly. 
With the prompt engineering template, as shown in Figure~\ref{fig: relation_wise_comparison_ALL}, ChatGPT does well on the \textit{Contrast}, \textit{Condition}, and \textit{Instantiation} second-level class. 
Appending the input-output example from each discourse relation as the prefix part of the prompt template helps solve this task easily.
Finally, the performance of ChatGPT on all second-level classes increases significantly except the \textit{Exp.List} subclass.

\begin{table}[!t]
\small
\centering
\setlength\tabcolsep{4pt}
\scalebox{0.85}{
\begin{tabular}{@{}l|cc|cc@{}}
\toprule
\multicolumn{1}{c|}{\multirow{2}{*}{\textbf{Method}}} & \multicolumn{2}{c|}{\textbf{Top}} & \multicolumn{2}{c}{\textbf{Second}} \\
& F1 & Acc & F1 & Acc \\
\midrule
Random & 24.74 & 25.47 & 6.48 & 8.78 \\
\citet{DBLP:conf/ijcai/LiuOSJ20} & 63.39 & 69.06 & 35.25 & 58.13\\
\citet{DBLP:journals/corr/abs-2211-13873} & 65.76 & 72.52 & 41.74 & 61.16\\
\citet{DBLP:conf/emnlp/LongW22} & 69.60 & 72.18 & 49.66 & 61.69\\
\citet{DBLP:journals/corr/abs-2305-03973}& 70.84 & 75.65 & 49.03 & 64.58\\
\hline
ChatGPT$_{\text{Prompt}}$ & 29.85 & 32.89 & 9.27 & 15.59 \\
ChatGPT$_{\text{PE}}$ & 33.78 & 34.94 & 10.73 & 20.31 \\
ChatGPT$_{\text{ICL}}$ & 36.11 & 44.18 & 16.20 & 24.54 \\
\bottomrule
\end{tabular}
}
\caption{
The performance of ChatGPT performs on the implicit discourse relation recognition task of PDTB (\textit{Ji}) test set. 
}
\label{tab:ChatGPT_IMP_Performance}
\end{table}

\subsubsection{Implicit Discourse Relation Recognition}
\label{sec:im-pdtb}
\paragraph{Experimental Results.} The performance in Table~\ref{tab:ChatGPT_IMP_Performance} demonstrates that \textbf{implicit discourse relation remains a challenging task for ChatGPT.} 
Even when using the information of label dependence and representative discourse connectives in the in-context learning setting, ChatGPT only achieves 24.54\% test accuracy and 16.20\% F1 score on the 11 second-level class of discourse relations. 
In particular, ChatGPT performs poorly on the second-level classes such as \textit{Comp.Concession}, \textit{Cont.Pragmatic Cause}, \textit{Exp.Alternative}, and \textit{Temp.Synchrony}. 
This may be because ChatGPT cannot understand the abstract sense of each discourse relation and the features from the text. 
When ChatGPT cannot capture the label sense and linguistic traits, it sometimes responds, "There doesn't appear to be a clear discourse relation between Argument 1 and Argument 2." or predicts as \textit{Cont.Cause} class.

\begin{figure*}[!t]
\centering
\includegraphics[width=\textwidth]{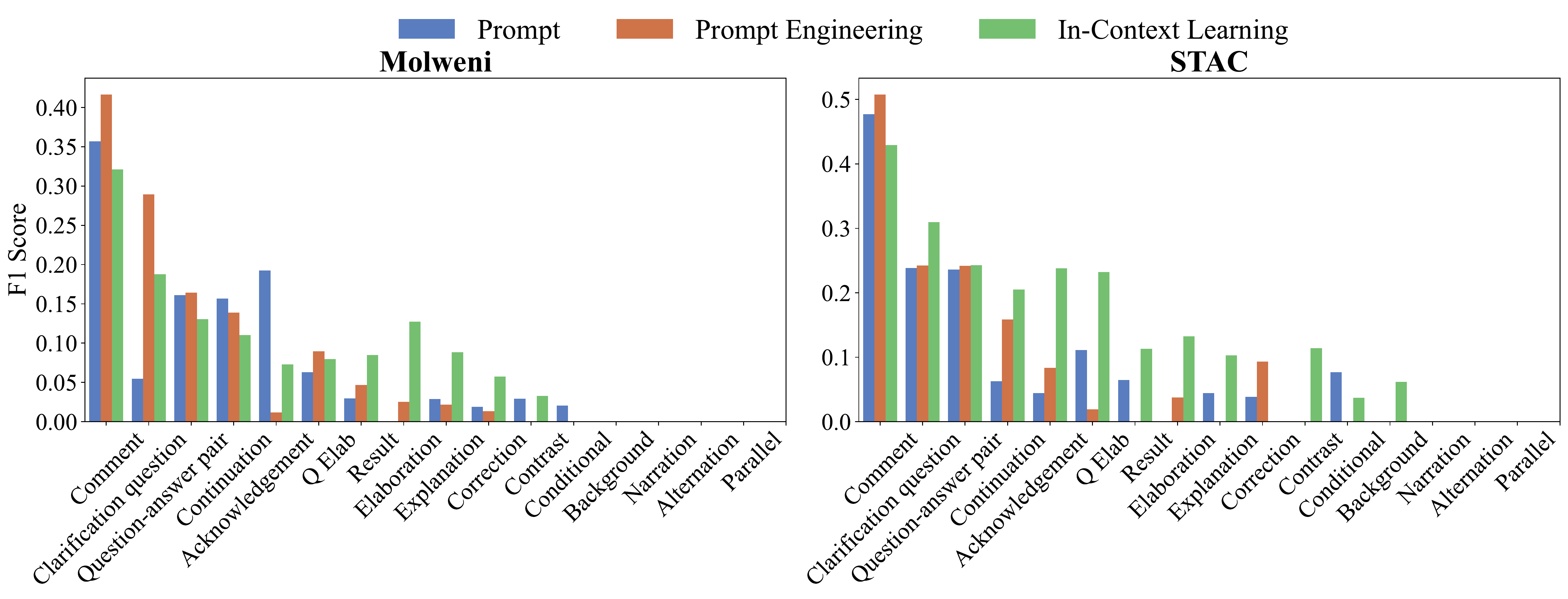}
\vspace{-0.3in}
\caption{Relation-wise performance comparison on dialogue benchmarks by ChatGPT with different prompting methods.}
\label{fig: relation_wise_comparison_dialogue}
\vspace{-0.6cm}
\end{figure*}
\subsection{Multi-genre Crowd-sourced Discourse Relation Recognition}

\paragraph{Detailed Experimental Setting.}



In this section, we evaluate the model on DiscoGeM \cite{scholman2022discogem}, which is a multi-genre implicit discourse relations dataset (details in Appendix~\ref{sec:experimental setting}).
For a fair and comprehensive evaluation, we test ChatGPT on the full test set containing 1,286 instances under the single label setting.
To help ChatGPT understand the relations, we verbalize the relations in different settings\footnote{We remove around 10 items with the ``differentcon'' relation as we do not find its explanation either in the paper or in the PDTB annotation guideline.}.
In addition to the vanilla setting where the model directly predicts labels (ChatGPT{\tiny Prompt}), we also replace relations that have special tokens or abbreviations with plain text, e.g. (``arg1-as-subst'' is replaced with ``argument 1 as substitution'').
Under this setting (ChatGPT{\tiny PE}), we concatenate the most typical connective\footnote{\url{https://github.com/merelscholman/DiscoGeM/blob/main/Appendix/DiscoGeM_ConnectiveMap.pdf}} to ChatGPT{\tiny Prompt}.
We further explored in-context learning (ChatGPT{\tiny ICL}):
We randomly sample 1 or 3 examples from the training set as demonstrations (ChatGPT{\tiny ICL-1} and ChatGPT{\tiny ICL-3}).
Following the setting in Section \ref{sec:im-pdtb}, we manually curated a set of 18 typical examples from the training dataset for each relation as demonstrations (ChatGPT{\tiny ICL-18}).
    


\paragraph{Experimental Results.}
\begin{table}[!t]
\small
\centering
\setlength\tabcolsep{1.5pt}

\scalebox{0.85}{\begin{tabular}{l|cc|cc|cc|cc}
\toprule
\multicolumn{1}{c|}{\multirow{2}{*}{\textbf{Method}}} & \multicolumn{2}{c|}{\textbf{All}} & \multicolumn{2}{c|}{\textbf{Europarl}} & \multicolumn{2}{c|}{\textbf{Novel}} & \multicolumn{2}{c}{\textbf{Wiki.}} \\
& Acc & F1 & Acc & F1& Acc & F1& Acc & F1 \\
\midrule
Random & 5.5 & 3.2 & 5.5 & 3.2 & 5.8 & 3.1 & 5.6 & 3.2\\
\cite{DBLP:conf/ijcai/LiuOSJ20} & 48.7 & 22.3 & 53.3 & 25.9 & 45.3 & 23.1 & 45.6 & 24.0 \\
\hline
ChatGPT\tiny{Prompt} & 10.8 & 3.5 & 13.7 & 4.2 & 9.9 & 3.7 & 9.4 & 3.1 \\
ChatGPT\tiny{PE} & 20.8 & 4.2 & 21.6 & 5.0 & 25.3 & 4.8 & 17.7 & 3.7 \\
ChatGPT\tiny{ICL-1} & 3.7 & 4.5 & 4.8 & 6.5 & 3.1 & 3.5 & 3.4 & 4.2\\
ChatGPT\tiny{ICL-3} & 3.3 & 2.8 & 3.1 & 2.4 & 4.3 & 4.2 & 2.9 & 2.5\\
ChatGPT\tiny{ICL-18} & 2.0 & 2.1 & 1.2 & 2.9 & 3.1 & 1.7 & 1.9 & 2.0\\
\bottomrule
\end{tabular}
}
\caption{Evaluation results (accuracy and Macro-averaged F1 score $\%$) on the DiscoGeM dataset. In addition to the performance on the full test set (``All''), we also report the genre-wise performance on different sub-sets (``Europarl'', ``Novel'', and ``Wiki.''). }
\label{tab:multigenre-results}
\vspace{-0.4cm}
\end{table}
Results are shown in Table \ref{tab:multigenre-results}.
We report performance from both the random baseline and the model \cite{DBLP:conf/ijcai/LiuOSJ20} fine-tuned on DiscoGeM (results reported in \cite{yung2022label}). Generally, while ChatGPT slightly outperforms the random baseline, it lags behind the supervised model \cite{DBLP:conf/ijcai/LiuOSJ20} by a significant margin (up to 30\% accuracy and 20\% macro-F1). 
Prompt engineering (ChatGPT{\tiny PE}) could improve ChatGPT's performance, possibly due to the introduction of verbalization of labels that provided additional information for task understanding.

However, the introduction of different kinds of in-context learning templates (ChatGPT{\tiny ICL}) did not have a positive influence on the model's ability to understand the task. 
In fact, the ChatGPT{\tiny ICL} model performed near-random or worse than random as the number of examples increased.
This is possibly due to the fact that implicit discourse relations can express more than one meaning \cite{rohde2016filling, scholman2017examples}, which makes it difficult to select representative and informative demonstrations. 
Overall, these findings suggest that it may require additional improvements or prompt engineering for ChatGPT to effectively perform tasks with complex classification requirements.

\begin{table}[!t]
\small
\centering
\setlength\tabcolsep{2pt}
\begin{tabular}{l|cc|cc}
\toprule
\multicolumn{1}{c|}{\multirow{2}{*}{\textbf{Method}}} & \multicolumn{2}{c|}{\textbf{STAC}} & \multicolumn{2}{c}{\textbf{Molweni}} \\
& Link & Link\&Rel & Link & Link\&Rel \\
\midrule
\citet{afantenos2015discourse} & 68.8 & 50.4 & - & - \\ 
\citet{perret2016integer} & 68.6 & 52.1 & - & - \\ 
\citet{shi2019deep} & 73.2 & 55.7 & 78.1 & 54.8 \\ 
\hline
ChatGPT\tiny{zero w/ desc.} & 20.5 & 4.3 & 26.7 & 5.0\\
ChatGPT\tiny{zero w/o desc.} & 20.0 & 4.4 & 28.3 & 5.4 \\
\hline
ChatGPT\tiny{few (n=1) w/ desc.} & 21.0 & 7.1 & 25.7 & 6.0\\
ChatGPT\tiny{few (n=3) w/ desc.} & 20.7 & 7.3 & 25.1 & 5.7\\
ChatGPT\tiny{few (n=1) w/o desc.} & 21.2 & 6.2 & 27.2 & 6.8 \\
ChatGPT\tiny{few (n=3) w/o desc.} & 21.3 & 7.4 & 26.5 & 6.9\\
\bottomrule
\end{tabular}
\vspace{-0.1in}
\caption{Evaluation results (Micro-averaged F1 score $\%$ on the multi-party dialogue parsing datasets STAC and Molweni. Both the zero- (ChatGPT{\tiny zero}) and few-shot (ChatGPT{\tiny few}) baselines are tested. Under each setting, there are two variants: whether to provide a description to the labels (w/ desc.) or not (w/o desc.). The label descriptions are from \citet{asher2016discourse}.}
\vspace{-0.6cm}
\label{tab:DialogueDP_results}
\end{table}

\vspace{-0.2cm}
\subsection{Dialogue Discourse Parsing}
The dialogue discourse parsing task \cite{asher2016discourse, shi2019deep} is proposed to evaluate the ability to understand and respond to multi-party conversations in a coherent and context-aware manner.
It focuses on extracting meaningful information from dialogues.
The goal of dialogue discourse parsing is to automatically identify the structural and semantic relationships among utterances, speakers, and topics in a conversation.

\paragraph{Detailed Experimental Setting.}
The setting of discourse parsing in multi-party dialogue can be formulated as follows.
Given a multi-party chat dialogue $D=\{u_1, u_2, ..., u_n\}$ with $n$ utterances ($u_1$ to $u_n$), a system is required to predict a graph $G(V, E, R)$, where $V$ is the vertex set containing all the utterances, $E$ is the predicted edge set between utterances, and $R$ is the predicted discourse relation set.
According to the content of outputs, there are three evaluation settings:
\begin{itemize}
    \item \textbf{Link prediction:} Given $D$, predict the links between utterances ($E$).
    Under this setting, the types of relations are ignored, and we only evaluate whether links are correctly predicted or not.
    \item \textbf{Link \& Relation prediction:} Given $D$, predict the links between utterances and classify the discourse relation for the predicted links ($E$ and $R$).
    Here, a true prediction requires both correctly predicting the link and its type of relation.
    \item \textbf{Relation classification:} Apart from the above two link prediction settings, we additionally evaluate ChatGPT's relation classification ability. 
    Here, the model is given $D$, and the ground truth links $E$, and is required to predict the corresponding relations $R$.
\end{itemize}

In this work, we evaluate ChatGPT's performance on two multi-party dialogue discourse parsing benchmarks: STAC 
\cite{asher2016discourse} and Molweni \cite{li2020molweni}.
Details are presented in Appendix~\ref{sec:experimental setting}.
    



\begin{table}[!t]
\small
\centering
\setlength\tabcolsep{4pt}

\begin{tabular}{l|cc|cc}
\toprule
\multicolumn{1}{c|}{\multirow{2}{*}{\textbf{Method}}} & \multicolumn{2}{c|}{\textbf{STAC}} & \multicolumn{2}{c}{\textbf{Molweni}} \\
& Acc & F1 & Acc & F1 \\
\midrule
Random & 6.2 & 4.8 & 6.3 & 4.1 \\
\hline
ChatGPT\tiny{Prompt} & 22.8 & 8.7 & 16.5 & 6.9 \\
ChatGPT\tiny{PE} & 25.9 & 8.6 & 23.0 & 7.6\\
ChatGPT\tiny{ICL} & 24.1 & 13.9 & 14.7 & 8.1\\
\bottomrule
\end{tabular}
\vspace{-0.1in}
\caption{Evaluation results (Accuracy and Macro-averaged F1 (\%)) on the multi-party dialogue parsing datasets STAC and Molweni. Here, the ChatGPT{\tiny Prompt}, ChatGPT{\tiny PE}, and ChatGPT{\tiny ICL} correspond to ChatGPT{\tiny zero w/o desc.}, ChatGPT{\tiny zero w/ desc.}, and ChatGPT{\tiny few (n=1) w/ desc.}, respectively. The relation-wise performance is visualized in Figure \ref{fig: relation_wise_comparison_dialogue}.}
\label{tab:DialogueDP_results-rel-only}
\vspace{-0.5cm}
\end{table}

\paragraph{Experimental Results.}
The evaluation results on the ``Link prediction'' and ``Link \& Relation prediction'' settings are presented in Table \ref{tab:DialogueDP_results}. 
ChatGPT performs significantly worse than the supervised baselines \cite{afantenos2015discourse, perret2016integer, shi2019deep} on both the link prediction and the link \& relation prediction settings.
Notably, on the link prediction setting, ChatGPT underperforms other baselines by up to 50\% F1.
It fails to give potential relations between utterances, indicating its poor understanding of the structure of multi-party dialogues.
Adding additional examples seems to improve ChatGPT's performance under the Link \& Relation prediction setting.
However, these examples could have an adverse effect on link prediction (e.g., on Molweni).
We also noticed that adding label descriptions does not help ChatGPT understand the task setting.
We present results under the ``Relation classification'' setting in Table \ref{tab:DialogueDP_results-rel-only}.
ChatGPT also does not achieve very high performance under this setting, which indicates the difficulty in understanding discourse relations in dialogues.
To sum up, \textbf{ChatGPT still suffers from a poor understanding of the dialogue structures} in multi-party dialogues and providing appropriate classifications.




\section{Conclusion and Future Work}
In conclusion, this study thoroughly examines ChatGPT's ability to handle pair-wise temporal relations, causal relations, and discourse relations by assessing its performance on the complete test sets of over 11 datasets. The result exhibits that even though ChatGPT obtains impressive zero-shot performance across other various tasks, there is still a gap for ChatGPT to achieve excellent performance on temporal and discourse relations. Though there may be numerous other capabilities of ChatGPT that go unnoticed in this paper, future work should nonetheless investigate the capability of ChatGPT on more tasks (e.g., analogy relation between two sentences~\cite{DBLP:journals/corr/abs-2310-12874}).


\section*{Limitation}
\paragraph{Evaluation Metrics}
In this paper, we exclusively assess the performance of ChatGPT on well-used evaluation metrics such as accuracy and F1 score. Nevertheless, these metrics are nonlinear or discontinuous metrics, and a recent study has revealed that such metrics yield conspicuous emergent capabilities, whereas linear or continuous metrics result in smooth, continuous predictable changes in model performance~\cite{DBLP:journals/corr/abs-2304-15004}. We intend to incorporate this aspect in forthcoming research endeavors.

\paragraph{Empirical Conclusions}
In this study, we give comprehensive comparisons and discussions of ChatGPT and prompts. All the conclusions are proposed based upon empirical analysis of the performance of ChatGPT to academic benchmarks. In light of the rapid evolution of the field, we will update the latest opinions timely.

\section*{Ethics Statement}
In this work, we conformed to accepted privacy practices and strictly followed the data usage policy. All evaluated dataset of this paper is publicly available, and this work is in the intended use. Since we do not introduce social and ethical bias into the model or amplify any bias from the data, we can foresee no direct social consequences or ethical issues.
Moreover, this study mainly formulates these sentence-level relations tasks as multi-choice tasks and requires ChatGPT to generate the English letter (e.g., "A," "B," "C," and "D"). Therefore, we do not observe or anticipate any potential toxicity, biases, or privacy in the generated context from ChatGPT. 
Furthermore, we also try our best to reduce these potential risks to prevent generating toxicity, biases, or privacy text by manually tailored prompt templates. These prompt templates only instruct ChatGPT to select the answer without any explanation.

\section*{Acknowledgements}
The authors of this paper were supported by the NSFC Fund (U20B2053) from the NSFC of China, the RIF (R6020-19 and R6021-20) and the GRF (16211520 and 16205322) from RGC of Hong Kong. We also thank the support from the UGC Research Matching Grants (RMGS20EG01-D, RMGS20CR11, RMGS20CR12, RMGS20EG19, RMGS20EG21, RMGS23CR05, RMGS23EG08).
\newpage
\bibliography{anthology,custom}
\bibliographystyle{acl_natbib}

\clearpage
\appendix
\section{Experimental Setting}\label{sec:experimental setting}
\subsection{Evaluation Dataset} \label{sec:appendix_statistics}
\paragraph{TB-Dense.}
TB-Dense \cite{cassidy2014tbdense} is a densely annotated dataset from TimeBank and TempEval \cite{uzzaman2013semeval} that contains six label types, including \textit{BEFORE}, \textit{AFTER}, \textit{SIMULTANEOUS}, \textit{NONE}, \textit{INCLUDES} and \textit{IS\_INCLUDED}.

\paragraph{MATRES.}
MATRES \cite{ning2018matres} is an annotated dataset that includes refined annotations from TimeBank \cite{pustejovsky2003timebank}, AQUAINT, and Platinum documents.
Four relations are annotated for the
start time comparison of event pairs in 275 documents, namely \textit{BEFORE}, \textit{AFTER}, \textit{EQUAL}, and \textit{VAGUE}. 
Note that the two relations named \textit{EQUAL} and \textit{VAGUE} are equivalent to \textit{SIMULTANEOUS} and \textit{NONE} in TB-Dense, respectively.

\paragraph{TDDMan.}
TDDMan is a subset of the TDDiscourse corpus \cite{naik2019tddiscourse}, which was created to explicitly emphasize global discourse-level temporal ordering.
Five temporal relations are annotated including \textit{BEFORE}, \textit{AFTER}, \textit{SIMULTANEOUS}, \textit{INCLUDES} and \textit{IS\_INCLUDED}.

\paragraph{COPA.}
The Choice of Plausible Alternatives (COPA)~\cite{DBLP:conf/semeval/GordonKR12} dataset is a collection of questions that require causality reasoning and inferences to solve.
Each question posits a commonly seen event, along with two possible options that either describe the \textit{cause} or \textit{effect} of the event.
This requires the model to identify the relationship between a cause and its effect and then select the most likely explanation for that relationship among a set of alternatives.
Such design makes COPA a very representative benchmark for evaluating causal relational reasoning.
In this paper, we use the testing split of COPA, consisting of 500 questions, for evaluation.

\paragraph{e-CARE.} 
The e-CARE~\cite{DBLP:conf/acl/DuDX0022} dataset is a large human-annotated commonsense causal reasoning benchmark that contains over 21,000 multiple-choice questions.
It is designed to provide a conceptual understanding of causality and includes free-text-formed conceptual explanations for each causal question to explain why the causation exists.
Each question either focuses on the \textit{cause} or \textit{effect} of a given event and consists of two possible explanations.
The model is still asked to select the more plausible one, given an event-and-relationship pair.
Since the testing set is not publicly available, we bank on 2,132 questions in the validation set for evaluating LLMs.

\paragraph{HeadlineCause.} 
HeadlineCause~\cite{DBLP:conf/lrec/GusevT22} is a dataset designed for detecting implicit causal relations between pairs of news headlines.
It includes over 5000 headline pairs from English news and over 9000 headline pairs from Russian news, labeled through crowd-sourcing. 
Given a pair of news, the model is first asked to determine whether a causal relationship exists between them.
If yes, it needs to further determine the role of cause and effect for the two news.
It serves as a very challenging and comprehensive benchmark for evaluating models' capability to detect causal relations in natural language text.
We select 542 English news pairs from the testing set that are used for evaluation.

\paragraph{The Penn Discourse Treebank 2.0 (PDTB 2.0).}
PDTB 2.0 is a large-scale corpus that comprises a vast collection of 2,312 articles from the Wall Street Journal (WSJ)~\cite{DBLP:conf/lrec/PrasadDLMRJW08}. It utilizes a lexically grounded approach to annotate discourse relations, with three sense levels (classes, types, and sub-types) naturally forming a natural sense hierarchy. In this dataset, we assess the performance of ChatGPT on a popular setting of the PDTB 2.0 dataset, known as the Ji-setting~\cite{DBLP:journals/tacl/JiE15}. This Ji-setting follows~\citet{DBLP:journals/tacl/JiE15} to divide sections 2-20, 0-1, and 21-22 into training, validation, and test sets, respectively. We evaluate ChatGPT on the whole test set of IDRR task and EDRR task with four top-level discourse relations (i.e., \textit{Comparison}, \textit{Contingency}, \textit{Expansion}, \textit{Temporal}) and the 11 major second-level discourse senses. The dataset statistics are displayed in Table~\ref{table:pdtb_statistics} and Table~\ref{table:pdtb_Sec_level_statistics} in Appendix.

\paragraph{DiscoGeM.}
The DiscoGeM dataset \cite{scholman2022discogem} is a crowd-sourced corpus of multi-genre implicit discourse relations.
Different from the expert-annotated PDTB, DiscoGeM adopts a crowd-sourcing method by asking crowd workers to provide possible \textit{connectives} between two arguments.
They curated a connective mapping from connectives to the discourse relation senses in PDTB, which is used to generate PDTB-style discourse relations from the crowd-sourced connectives.
Clear differences in the distributions across three genres have been observed \cite{scholman2022discogem}.
For instance, \textsc{conjunction} is more prevalent in Wikipedia text, and \textsc{precedence} occurs more frequently in novels than in other genres.
DiscoGeM includes 6,505 instances from three genres: political speech data from the Europarl corpus, texts from 20 novels, and encyclopedic texts from English Wikipedia.
The data was split into 70\% training, 20\% testing, and 10\% development sets.
For a fair and comprehensive evaluation, we test ChatGPT on the full test set containing 1,286 instances under the single label setting.

\paragraph{STAC} \cite{asher2016discourse}  was the first corpus of discourse parsing for multi-party dialogue. The dataset was adapted from an online multi-player game \textit{The Settlers of Catan}, where players acquire and trade resources in order to build facilities. The STAC corpus came from the chat history in trade negotiations.

\paragraph{Molweni} \cite{li2020molweni} came from the large-scale multi-party dialogue dataset, \textit{the Ubuntu Chat Corpus} \cite{lowe2015ubuntu}, which is a collection of chat logs between users seeking technical support on the Ubuntu operating system.\citet{li2020molweni} conducted additional annotations specific to dialogue discourse parsing to construct the Molweni dataset, which is larger in scale than STAC.
Moreover, a preliminary study on Molweni has shown comparable baseline performance to that in STAC, which indicates the two datasets have similar quality and complexity

\subsection{ChatGPT Hyperparameter}
In this study, we only call the OpenAI API for conducting evaluation and do not use any GPU to train the model. For the hyperparameter for ChatGPT response generation, the temperature is 0.7, Top\_p is 1, and the max\_tokens is 256.

\begin{table}[!t]
\small
\centering
\setlength\tabcolsep{4pt}
\begin{tabular}{l|c|c|c|c}
\toprule
\multicolumn{1}{c|}{\textbf{Dataset}}  & \textbf{Train} & \textbf{Validation} & \textbf{Test}  & \textbf{\# of labels} \\
\midrule
TB-Dense & 4,032 & 629 & 1,427 & 6            \\
MATRES   & 6,336 & $-$   & 837   & 4            \\
TDDMan   & 4,000 & 650 & 1,500 & 5    \\
\bottomrule
\end{tabular}
\caption{Statistics of three temporal relation datasets.}
\label{tab:temporal_statistics}
\end{table}

\begin{figure}[!t]
\small
\centering
\includegraphics[width=\linewidth]{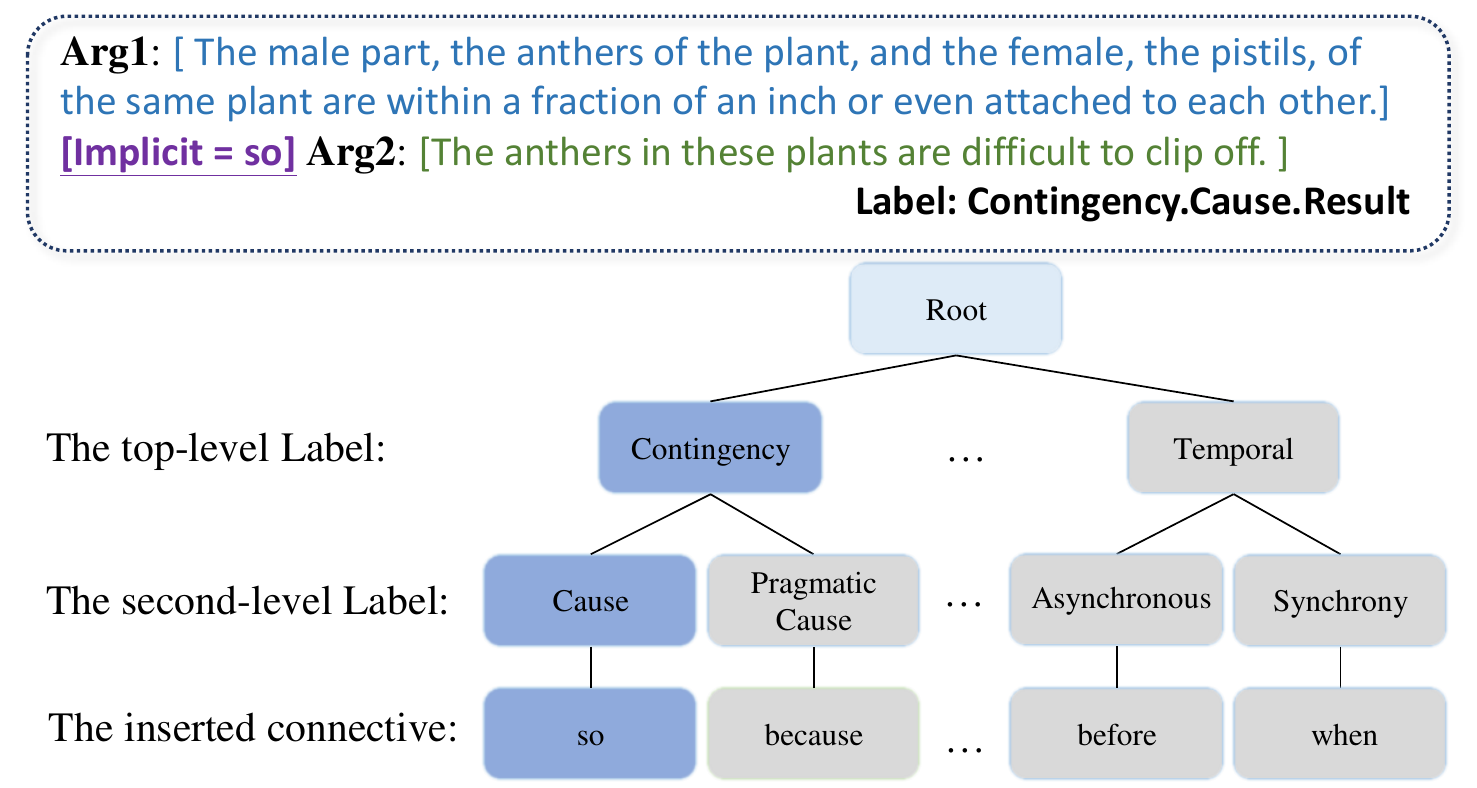}
\vspace{-0.2in}
\caption{An example of the implicate discourse relation recognition task and the label hierarchy.}
\label{fig:Implict discourse relation example}
\vspace{-0.1in}
\end{figure}

\begin{figure}[!t]
    \centering
    \includegraphics[width=\linewidth]{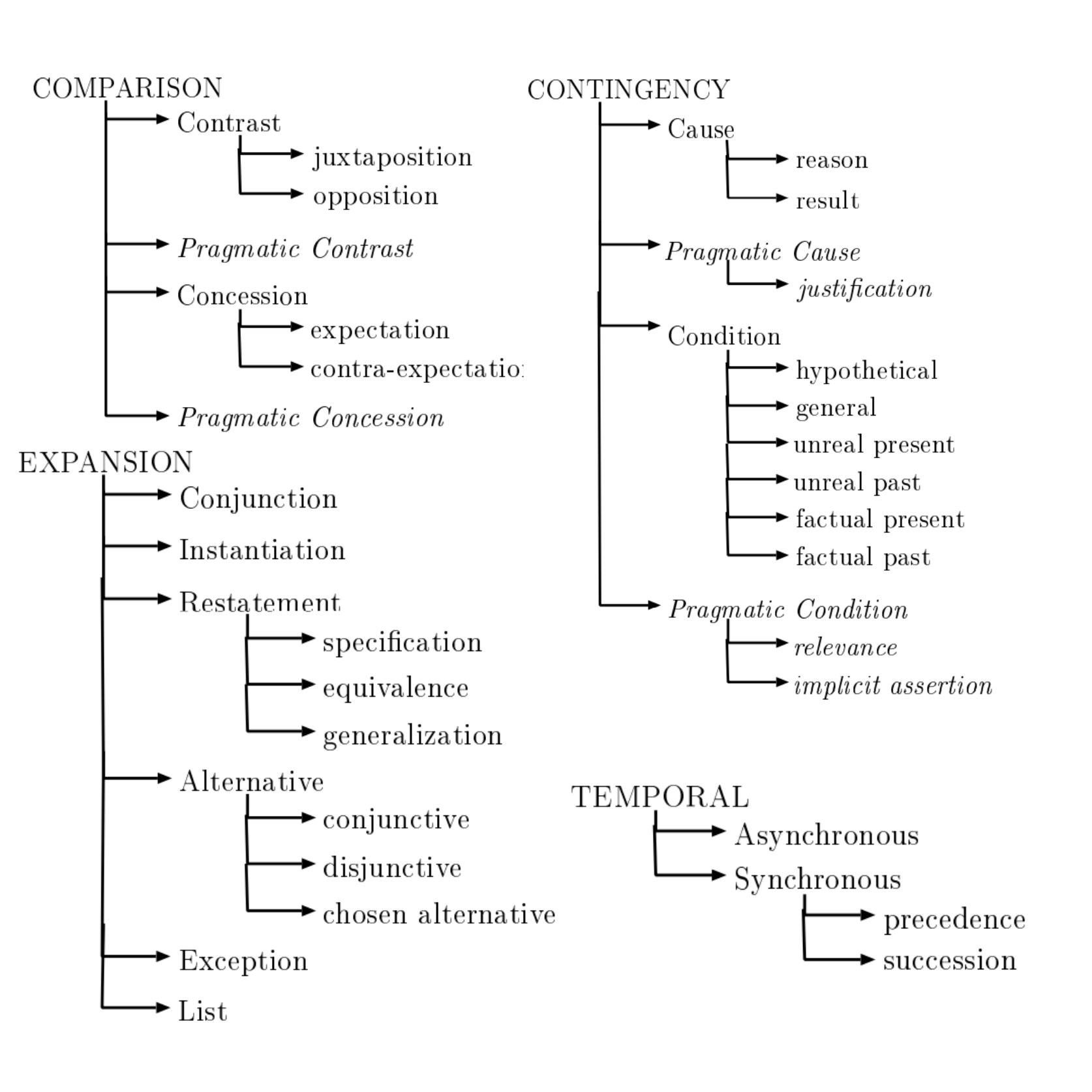}
    \vspace{0.1in}
    \caption{The sense hierarchy of implicit discourse relation in PDTB2.0 dataset}
    \label{fig:PDTB_Senses_Hierarchy}
\end{figure}

\begin{table}[!t]
\small
\centering
\begin{tabular}{l|c|c|c}
\toprule
\textbf{Top-level Senses}  & \textbf{Train} & \textbf{Validation} & \textbf{Test} \\
\midrule
Comparison & 1,942 & 197 & 152 \\
Contingency & 3,342 & 295 & 279 \\
Expansion & 7,004 & 671 & 574 \\
Temporal & 760 & 64 & 85 \\
\hline
Total & 12,362 & 1,183 & 1,046\\
\bottomrule
\end{tabular}
\caption{Statistics of four top-level implicit senses in PDTB 2.0.}
\label{table:pdtb_statistics}
\end{table}

\begin{table}[!t]
\small
\centering
\begin{tabular}{l|c|c|c}
\toprule
\textbf{Second-level Senses}  & \textbf{Train} & \textbf{Validation} & \textbf{Test} \\
\midrule
Comp.Concession & 180 & 15 & 17\\
Comp.Contrast & 1566 & 166 & 128\\
Cont.Cause & 3227 & 281 & 269\\
Cont.Pragmatic Cause & 51 & 6 & 7\\
Exp.Alternative & 146 & 10 & 9\\
Exp.Conjunction & 2805 & 258 & 200\\
Exp.Instantiation & 1061 & 106 & 118\\
Exp.List & 330 & 9 & 12\\
Exp.Restatement & 2376 & 260 & 211\\
Temp.Asynchronous & 517 & 46 & 54\\
Temp.Synchrony & 147 & 8 & 14\\
\hline
Total &  12406 & 1165 & 1039\\
\bottomrule
\end{tabular}
\caption{The implicit discourse relation data statistics of second-level types in PDTB 2.0.}
\label{table:pdtb_Sec_level_statistics}
\end{table}

\begin{table}[!t]
\small
\centering
\setlength\tabcolsep{1.5pt}
\begin{tabular}{l|l|c}
\toprule
\multicolumn{1}{c|}{\textbf{Dataset}} & \multicolumn{1}{c|}{\textbf{Data source}} & \multicolumn{1}{c}{\textbf{\# of dialogues/utterances/relations}} \\
\midrule
STAC & \specialcell{Online multi-\\player game} & \specialcell{111 \\ 1156 \\ 1128} \\
\hline
Molweni & \specialcell{The Ubuntu \\ chat corpus} & \specialcell{500 \\ 4430 \\ 3911} \\
\bottomrule
\end{tabular}
\caption{Statistics of the multi-party dialogue parsing datasets STAC and Molweni.}
\label{tab:DialogueDP_datasets}
\end{table}

\section{Downstream Tasks of Discourse Relations}\label{sec:appendix_downstreamtask}

Discourse relations can be applied for acquiring commonsense knowledge and developing discourse-aware sophisticated commonsense reasoning benchmarks that are shown to be hard for current large language models~\cite{DBLP:conf/emnlp/Bhargava022DISCOSENSE}. 
In this section, we study two NLP tasks that are applications of discourse relations, one for commonsense acquisition~\cite{DBLP:conf/emnlp/FangWCHZSH21CKBP,DBLP:journals/corr/abs-2304-10392} 
and one for a commonsense question answering constructed with sophisticated discourse markers~\cite{DBLP:conf/emnlp/Bhargava022DISCOSENSE}.

\paragraph{Commonsense Knowledge Base Population.}~CKBP~\cite{DBLP:conf/emnlp/FangWCHZSH21CKBP} is a benchmark for populating commonsense knowledge from discourse knowledge triples. For example, it requires the model to determine whether a discourse knowledge entry (\textit{John drinks coffee}, \texttt{Succession/then}, \textit{John feels refreshed}) represents a plausible commonsense knowledge, (\textit{PersonX drinks coffee}, \texttt{xReact}, \textit{refreshed}), a form of social commonsense knowledge defined in ATOMIC~\cite{DBLP:conf/aaai/SapBABLRRSC19} where \texttt{xReact} studies what would \textit{PersonX} feels after the head event.
We include the latest test set of CKBP v2\footnote{\url{https://github.com/HKUST-KnowComp/CSKB-Population/}} for our experiments, which contains 4k triples converted from discourse relations to 15 commonsense relations defined in ConceptNet~\cite{DBLP:conf/aaai/SpeerCH17}, ATOMIC~\cite{DBLP:conf/aaai/SapBABLRRSC19}, and GLUCOSE~\cite{DBLP:conf/emnlp/MostafazadehKMB20}. Prompt templates are presented in Table~\ref{tab:ckbp_pemplate}.

\paragraph{\textsc{DiscoSense}.} \textsc{DiscoSense} is a commonsense question-answering dataset built upon discourse connectives.
It's constructed from \textsc{Discovery}~\cite{DBLP:conf/naacl/SileoCPM19} and \textsc{Discofuse}~\cite{DBLP:conf/naacl/GevaMSB19} where there are two sentences connected through a discourse connective and the negative options are generated through a conditional adversarial filtering process to make sure the difficulty of the dataset. The task is defined as selecting the most plausible coming sentence given the source sentence and a discourse connective such as \textit{because}, \textit{although}, \textit{for example}, etc.
Supervised learning models struggle on this dataset, showing a lack of subtle reasoning ability for discourse relations. We take the test set for evaluation. Prompt templates are presented in Table~\ref{tab:discosense_template}.

\begin{table}[t]
\small
\centering

\begin{tabular}{l|cc|c}
\toprule
\multicolumn{1}{c|}{\multirow{2}{*}{\textbf{Method}}} & \multicolumn{2}{c|}{\textbf{CKBP v2.}} & \multicolumn{1}{c}{\textsc{\textbf{DiscoSense}}} \\
 & AUC & F1 & Acc \\
\midrule
Fine-tuned SOTA & 73.70 & 46.70 & 65.87 \\
ChatGPT$_{\text{PE}}$ & 65.77 & 45.93 & 47.25 \\
ChatGPT$_{\text{ICL}}$ & 66.20 & 46.42 & 54.67 \\
\bottomrule
\end{tabular}
\caption{
Performance on CSKB Population and \textsc{DiscoSense}.PE and ICL indicate the prompt engineering template and in-context learning prompt template.
}
\label{tab:downstream_application}
\end{table}

\paragraph{Experimental Results.} We present the experimental results on Table~\ref{tab:downstream_application}. We compare the performance of zero-shot ChatGPT with supervised SOTA, which is PseudoReasoner-RoBERTa-large~\cite{DBLP:conf/emnlp/FangDZSWS22} for CKBP v2 and Electra-large~\cite{DBLP:conf/iclr/ClarkLLM20} for \textsc{DiscoSense}. ChatGPT can achieve comparable F1 scores for CKBP v2. while still down performs regarding AUC. For the \textsc{DiscoSense} dataset, ChatGPT has a long way to reaching fine-tuned SOTA, letting alone human performance, indicating a lack of subtle reasoning ability to distinguish different discourse relations.

We report our experimental results summarized in Table~\ref{tab:downstream_application} leveraging the full test sets of both CKBP and \textsc{DiscoSense}. 
We compare the performance of zero-shot ChatGPT with that of PseudoReasoner-RoBERTa-large~\cite{DBLP:conf/emnlp/FangDZSWS22} for CKBP v2 and ELECTRA-large~\cite{DBLP:conf/iclr/ClarkLLM20} for \textsc{DiscoSense}, both of which are supervised state-of-the-arts. 
Our results show that ChatGPT achieves comparable F1 scores for CKBP v2, but it still underperforms in terms of AUC. 
For the \textsc{DiscoSense} dataset, ChatGPT has a long way to go to match the fine-tuned state-of-the-art performance, let alone human performance (95.40). 
This suggests that ChatGPT still lacks the subtle reasoning ability needed to distinguish between different discourse relations for making inferences.


\section{Prompt Templates} \label{sec:appendix_prompts}
The prompting or prompt tuning method is widely applied for many downstream tasks in the Natural Language Processing (NLP) field, the sensitivity and performance variance of the prompting method has been reported in a lot of works~\cite{han2021ptr,DBLP:journals/corr/abs-2309-08303, zhong2021factual, liu2021gpt, DBLP:conf/emnlp/LiGFXHMS23, DBLP:conf/icmlc2/ChanC23}. Therefore, we utilized the expert knowledge on these sentence-level relation classification tasks to manually craft a prompt template that outperformed a baseline~\cite{robinson2022leveraging} with fairly standard settings for all tasks. Our designed prompt template will be comprehensive and reliable baselines to exclude the variance of the prompt engineering and offer fair comparison baselines for further works. 
We list all prompt templates used in this paper as follows.


\begin{table*}
\small
\centering
\begin{tabular}{c|c|c|c|c}
\toprule
\multicolumn{5}{c}{\textbf{TB-Dense}}                                                   \\
\midrule
\multicolumn{1}{c|}{\textbf{Strategies}} & \textbf{Template input} & \textbf{ChatGPT} & \textbf{Gold} & \textbf{T/F} \\
\midrule
\specialcell{Prompt}              & \parbox[c]{8.5cm}{\texttt{Sentence:} The Organization of African Unity said Friday it would investigate the Hutu-organized genocide of more than 500,000 minority Tutsis in Rwanda nearly four years ago. Foreign ministers of member-states meeting in the Ethiopian capital agreed to set up a seven-member panel to investigate who shot down Rwandan President Juvenal Habyarimana's plane on April 6, 1994.\\\texttt{event1:} investigate\\\texttt{event2:} shot\\\texttt{Question:} What is the temporal relation between event1 and event2 in the sentence?\\\texttt{A. AFTER}\\\texttt{B. BEFORE}\\\texttt{C. SIMULTANEOUS}\\\texttt{D. NONE}\\\texttt{E. INCLUDES}\\\texttt{F. IS\_INCLUDED}\\\texttt{Answer:}}                       & NONE                   & AFTER         & F            \\
\midrule
\specialcell{Prompt\\Engineering}  & \parbox[c]{8.5cm}{\texttt{Determine the temporal order from "}investigate\texttt{" to "}shot\texttt{" in the following sentence: "}The Organization of African Unity said Friday it would investigate the Hutu-organized genocide of more than 500,000 minority Tutsis in Rwanda nearly four years ago. Foreign ministers of member-states meeting in the Ethiopian capital agreed to set up a seven-member panel to investigate who shot down Rwandan President Juvenal Habyarimana's plane on April 6, 1994.\texttt{". Only answer one word from AFTER, BEFORE, SIMULTANEOUS, NONE, INCLUDES, IS\_INCLUDED. Answer:}}                       & AFTER                   & AFTER         & T            \\
\bottomrule
\end{tabular}
\caption{Prompt example for TB-Dense.}
\label{tab:temporal_TB-Dense_templete1}
\end{table*}

\begin{table*}
\small
\centering
\begin{tabular}{c|c|c|c|c}
\toprule
\multicolumn{5}{c}{\textbf{TB-Dense}}                                                   \\
\midrule
\multicolumn{1}{c|}{\textbf{Strategies}} & \textbf{Template input} & \textbf{ChatGPT} & \textbf{Gold} & \textbf{T/F} \\
\midrule
\specialcell{In-Context\\Learning} & \parbox[c]{8.5cm}{\texttt{Determine the temporal order from "convictions" to "fraud" in the following sentence: "A federal appeals court has reinstated his state convictions for securities fraud.". Only answer one word from AFTER, BEFORE, SIMULTANEOUS, NONE, INCLUDES, IS\_INCLUDED. Answer: AFTER\\\\Determine the temporal order from "arrested" to "said" in the following sentence: "Derek Glenn, a spokesman for the Newark Police Department, said that of nine women who had been killed last year, suspects had been arrested in only four cases.". Only answer one word from AFTER, BEFORE, SIMULTANEOUS, NONE, INCLUDES, IS\_INCLUDED. Answer: BEFORE\\\\Determine the temporal order from "assassination" to "touched" in the following sentence: "The assassination touched off a murderous rampage by Hutu security forces and civilians, who slaughtered mainly Tutsis but also Hutus who favored reconciliation with the minority.". Only answer one word from AFTER, BEFORE, SIMULTANEOUS, NONE, INCLUDES, IS\_INCLUDED. Answer: SIMULTANEOUS\\\\Determine the temporal order from "seen" to "created" in the following sentence: "\"I haven't seen a pattern yet,\" said Patricia Hurt, the Essex County prosecutor, who created the task force on Tuesday.". Only answer one word from AFTER, BEFORE, SIMULTANEOUS, NONE, INCLUDES, IS\_INCLUDED. Answer: NONE\\\\Determine the temporal order from "meeting" to "agreed" in the following sentence: "Foreign ministers of member\-states meeting in the Ethiopian capital agreed to set up a seven\-member panel to investigate who shot down Rwandan President Juvenal Habyarimana's plane on April 6, 1994.". Only answer one word from AFTER, BEFORE, SIMULTANEOUS, NONE, INCLUDES, IS\_INCLUDED. Answer: INCLUDES\\\\Determine the temporal order from "investigation" to "said" in the following sentence: "The panel will be based in Addis Ababa, and will finish its investigation within a year, it said.". Only answer one word from AFTER, BEFORE, SIMULTANEOUS, NONE, INCLUDES, IS\_INCLUDED. Answer: IS\_INCLUDED}\\\\\texttt{Determine the temporal order from "}investigate\texttt{" to "}shot\texttt{" in the following sentence: "}The Organization of African Unity said Friday it would investigate the Hutu-organized genocide of more than 500,000 minority Tutsis in Rwanda nearly four years ago. Foreign ministers of member-states meeting in the Ethiopian capital agreed to set up a seven-member panel to investigate who shot down Rwandan President Juvenal Habyarimana's plane on April 6, 1994.\texttt{". Only answer one word from AFTER, BEFORE, SIMULTANEOUS, NONE, INCLUDES, IS\_INCLUDED. Answer:}}                       & BEFORE                  & AFTER         & F     \\
\bottomrule
\end{tabular}
\caption{Prompt example for TB-Dense.}
\label{tab:temporal_TB-Dense_templete2}
\end{table*}

\begin{table*}
\small
\centering
\begin{tabular}{c|c|c|c|c}
\toprule
\multicolumn{5}{c}{\textbf{MATRES}}                                                   \\
\midrule
\multicolumn{1}{c|}{\textbf{Strategies}} & \textbf{Template input} & \textbf{ChatGPT} & \textbf{Gold} & \textbf{T/F} \\
\midrule
\specialcell{Prompt}              & \parbox[c]{8.5cm}{\texttt{Sentence:} "It had a multiplying effect.", "We were pleased that England and New Zealand knew about it, and we thought that's where it would stop."\\\texttt{event1:} had\\\texttt{event2:} pleased\\\texttt{Question:} What is the temporal relation between event1 and event2 in the sentence?\\\texttt{A. AFTER}\\\texttt{B. BEFORE}\\\texttt{C. EQUAL}\\\texttt{D. VAGUE}\\\texttt{Answer:}}                       & AFTER                   & EQUAL         & F            \\
\midrule
\specialcell{Prompt\\Engineering}  & \parbox[c]{8.5cm}{\texttt{Determine the temporal order from "}had\texttt{" to "}pleased\texttt{" in the following sentence: "}"It had a multiplying effect.", "We were pleased that England and New Zealand knew about it, and we thought that's where it would stop. "\texttt{". Only answer one word from AFTER, BEFORE, EQUAL, VAGUE. Answer:}}                       & EQUAL                   & EQUAL         & T            \\
\midrule
\specialcell{In-Context\\Learning} & \parbox[c]{8.5cm}{\texttt{Determine the temporal order from "give" to "tried" in the following sentence: "It will give the rest of the world the view that Cuba is like any other nation, something the US has, of course, tried to persuade the world that it is not.". Only answer one word from AFTER, BEFORE, EQUAL, VAGUE. Answer: AFTER\\\\Determine the temporal order from "invited" to "come" in the following sentence: "Fidel Castro invited John Paul to come for a reason.". Only answer one word from AFTER, BEFORE, EQUAL, VAGUE. Answer: BEFORE\\\\Determine the temporal order from "earned" to "rose" in the following sentence: "In the nine months, EDS earned \$315.8 million, or \$2.62 a share, up 13 \% from \$280.7 million, or \$2.30 a share.". Only answer one word from AFTER, BEFORE, EQUAL, VAGUE. Answer: EQUAL\\\\Determine the temporal order from "created" to "become" in the following sentence: "Ms. Atimadi says the war has created a nation of widows. Women have become the sole support of their families.". Only answer one word from AFTER, BEFORE, EQUAL, VAGUE. Answer: VAGUE}\\\\\texttt{Determine the temporal order from "}had\texttt{" to "}pleased\texttt{" in the following sentence: "}"It had a multiplying effect.", "We were pleased that England and New Zealand knew about it, and we thought that's where it would stop. "\texttt{". Only answer one word from AFTER, BEFORE, EQUAL, VAGUE. Answer:}}                       & BEFORE                  & EQUAL         & F     \\
\bottomrule
\end{tabular}
\caption{Prompt example for MATRES.}
\label{tab:temporal_MATRES_templete}
\end{table*}

\begin{table*}
\small
\centering
\begin{tabular}{c|c|c|c|c}
\toprule
\multicolumn{5}{c}{\textbf{TDDMan}}                                                   \\
\midrule
\multicolumn{1}{c|}{\textbf{Strategies}} & \textbf{Template input} & \textbf{ChatGPT} & \textbf{Gold} & \textbf{T/F} \\
\midrule
\specialcell{Prompt}              & \parbox[c]{8.5cm}{\texttt{Sentence:} The assassination touched off a murderous rampage by Hutu security forces and civilians, who slaughtered mainly Tutsis but also Hutus who favored reconciliation with the minority. It also reignited the civil war. The panel also will look at the exodus of about 2 million Rwanda Hutus to neighboring countries where they lived in U.N.-run refugee camps for 2 1/2 years.\\\texttt{event1:} rampage\\\texttt{event2:} exodus\\\texttt{Question:} What is the temporal relation between event1 and event2 in the sentence?\\\texttt{A. AFTER}\\\texttt{B. BEFORE}\\\texttt{C. SIMULTANEOUS}\\\texttt{D. INCLUDES}\\\texttt{E. IS\_INCLUDED}\\\texttt{Answer:}}                      & AFTER                   & BEFORE         & F            \\
\midrule
\specialcell{Prompt\\Engineering}  & \parbox[c]{8.5cm}{\texttt{Determine the temporal order from "}rampage\texttt{" to "}exodus\texttt{" in the following sentence: "}The assassination touched off a murderous rampage by Hutu security forces and civilians, who slaughtered mainly Tutsis but also Hutus who favored reconciliation with the minority. It also reignited the civil war. The panel also will look at the exodus of about 2 million Rwanda Hutus to neighboring countries where they lived in U.N.-run refugee camps for 2 1/2 years.\texttt{". Only answer one word from AFTER, BEFORE, SIMULTANEOUS, INCLUDES, IS\_INCLUDED. Answer:}}                       & BEFORE                   & BEFORE         & T            \\
\bottomrule
\end{tabular}
\caption{Prompt example for TDDMan.}
\label{tab:temporal_TDDMan_templete1}
\end{table*}

\begin{table*}
\small
\centering
\begin{tabular}{c|c|c|c|c}
\toprule
\multicolumn{5}{c}{\textbf{TDDMan}}                                                   \\
\midrule
\multicolumn{1}{c|}{\textbf{Strategies}} & \textbf{Template input} & \textbf{ChatGPT} & \textbf{Gold} & \textbf{T/F} \\
\midrule
\specialcell{In-Context\\Learning} & \parbox[c]{8.5cm}{\scriptsize\texttt{Determine the temporal order from "thrown" to "raised" in the following sentence: "Keating's convictions were thrown out in nineteen ninety-six on a technicality. And on that basis Keating was released from prison before he was eligible for parole. Now the ninth US circuit court of appeals has ruled that the original appeal was flawed since it brought up issues that had not been raised before.". Only answer one word from AFTER, BEFORE, SIMULTANEOUS, NONE, INCLUDES, IS\_INCLUDED. Answer: AFTER\\\\Determine the temporal order from "seized" to "parole" in the following sentence: "The bonds became worthless when the bankrupt thrift was seized by government regulators. Keating's convictions were thrown out in nineteen ninety-six on a technicality. And on that basis Keating was released from prison before he was eligible for parole.". Only answer one word from AFTER, BEFORE, SIMULTANEOUS, NONE, INCLUDES, IS\_INCLUDED. Answer: BEFORE\\\\Determine the temporal order from "assassination" to "reignited" in the following sentence: "The assassination touched off a murderous rampage by Hutu security forces and civilians, who slaughtered mainly Tutsis but also Hutus who favored reconciliation with the minority. It also reignited the civil war.". Only answer one word from AFTER, BEFORE, SIMULTANEOUS, NONE, INCLUDES, IS\_INCLUDED. Answer: SIMULTANEOUS\\\\Determine the temporal order from "war" to "genocide" in the following sentence: "It also reignited the civil war. The panel also will look at the exodus of about 2 million Rwanda Hutus to neighboring countries. The investigation will consider the role of \"internal and external forces prior to the genocide and subsequently, and the role of the United Nations and its agencies and the OAU before, during and after the genocide,\" the OAU said.". Only answer one word from AFTER, BEFORE, SIMULTANEOUS, NONE, INCLUDES, IS\_INCLUDED. Answer: INCLUDES\\\\Determine the temporal order from "arrests" to "related" in the following sentence: "But over all, arrests were made in more than 60 percent of murder cases, he said. Eight of the 14 killings since 1993 were already under investigation by the Newark Police Department, Glenn said. Of the eight victims, three were stabbed, two were strangled, two were beaten to death and one was asphyxiated, he said, and these different methods of killing and other evidence seem to indicate that the eight cases are not related.". Only answer one word from AFTER, BEFORE, SIMULTANEOUS, NONE, INCLUDES, IS\_INCLUDED. Answer: IS\_INCLUDED}\\\\\texttt{Determine the temporal order from "}rampage\texttt{" to "}exodus\texttt{" in the following sentence: "}The assassination touched off a murderous rampage by Hutu security forces and civilians, who slaughtered mainly Tutsis but also Hutus who favored reconciliation with the minority. It also reignited the civil war. The panel also will look at the exodus of about 2 million Rwanda Hutus to neighboring countries where they lived in U.N.-run refugee camps for 2 1/2 years.\texttt{". Only answer one word from AFTER, BEFORE, SIMULTANEOUS, INCLUDES, IS\_INCLUDED. Answer:}}                       & AFTER                  & BEFORE         & F     \\
\bottomrule
\end{tabular}
\caption{Prompt example for TDDMan.}
\label{tab:temporal_TDDMan_templete2}
\end{table*}

\begin{table*}
\small
\centering
\begin{tabular}{c|c|c|c|c}
\toprule
\multicolumn{5}{c}{\textbf{COPA}}  \\
\midrule
\multicolumn{1}{c|}{\textbf{Strategies}} & \textbf{Template input} & \textbf{ChatGPT} & \textbf{Gold} & \textbf{T/F} \\
\midrule
\specialcell{Prompt}  & \parbox[c]{9cm}{\texttt{The} cause \texttt{of} The cashier opened the cash register \texttt{is: 1.} The customer searched his wallet. \texttt{2.} The customer handed her money. \texttt{Only answer '1' or '2' only without any other words.}} & 2. & 2 & T \\
\midrule
\specialcell{Prompt\\Engineering}  & \parbox[c]{9cm}{\texttt{Given the event} The cashier opened the cash register\texttt{, which choice is more likely to be the} cause \texttt{of this event?\\1.} The customer searched his wallet.\\\texttt{2.} The customer handed her money. \\ \texttt{Only answer '1' or '2' only without any other words.}} & 2. & 2 & T \\
\midrule
\specialcell{In-Context\\Learning}  & \parbox[c]{9cm}{\texttt{Given the event The shirt shrunk, the cause of this event is likely to be I put it in the dryer. \\ Given the event It got dark outside, the effect of this event is likely to be The moon became visible in the sky. \\ Given the event} The cashier opened the cash register\texttt{, which choice is more likely to be the} cause \texttt{of this event? \\ 1.} The customer searched his wallet. \\ \texttt{2.} The customer handed her money. \\ \texttt{Only answer '1' or '2' only without any other words.}} & 2 & 2 & T \\
\bottomrule
\end{tabular}
\caption{Prompt templates used for the COPA benchmark.}
\label{tab:copa_pemplate}
\end{table*}

\begin{table*}
\small
\centering
\begin{tabular}{c|c|c|c|c}
\toprule
\multicolumn{5}{c}{\textbf{e-CARE}}                                                   \\
\midrule
\multicolumn{1}{c|}{\textbf{Strategies}} & \textbf{Template input} & \textbf{ChatGPT} & \textbf{Gold} & \textbf{T/F} \\
\midrule
\specialcell{Prompt}  & \parbox[c]{9cm}{\texttt{The} effect \texttt{of} They walked along the stream \texttt{is: 1.} They found lots of fish in it. \texttt{2.} They went to ponds. \texttt{Only answer '1' or '2' only without any other words.}} & 2. & 1 & F \\
\midrule
\specialcell{Prompt\\Engineering}  & \parbox[c]{9cm}{\texttt{Given the event} They walked along the stream\texttt{, which choice is more likely to be the} effect \texttt{of this event?\\1.} They found lots of fish in it. \\ \texttt{2.} They went to ponds. \\ \texttt{Only answer '1' or '2' only without any other words.}} & 1. & 1 & T \\
\midrule
\specialcell{In-Context\\Learning}  & \parbox[c]{9cm}{\texttt{Given the event There is a light rain today, the effect of this event is likely to be The roots of many plants are not moistened by rain. \\ Given the event His parents stopped him, the cause of this event is likely to be The child ran towards hippos. \\ Given the event} They walked along the stream\texttt{, which choice is more likely to be the} effect \texttt{of this event? \\ 1.} They found lots of fish in it. \\ \texttt{2.} They went to ponds. \\ \texttt{Only answer '1' or '2' only without any other words.}} & 1. & 1 & T \\
\bottomrule
\end{tabular}
\caption{Prompt templates used for the e-CARE benchmark.}
\label{tab:ecare_pemplate}
\end{table*}

\begin{table*}
\small
\centering
\begin{tabular}{c|c|c|c|c}
\toprule
\multicolumn{5}{c}{\textbf{HeadlineCause}}                                                   \\
\midrule
\multicolumn{1}{c|}{\textbf{Strategies}} & \textbf{Template input} & \textbf{ChatGPT} & \textbf{Gold} & \textbf{T/F} \\
\midrule
\specialcell{Prompt}  & \parbox[c]{9cm}{\texttt{News title A:} Guv encourages creative developers during lockdown. \texttt{News title B:} Govt hints at lockdown extension, but promises relaxations. \texttt{Is there any causal relationship between these two titles? 1. No. 2. A causes B. 3. B causes A. Only answer '1' or '2' or '3' without any other words.}} & 1 & 1 & T \\
\midrule
\specialcell{Prompt\\Engineering}  & \parbox[c]{9cm}{\texttt{News title A:} Guv encourages creative developers during lockdown. \texttt{News title B:} Govt hints at lockdown extension, but promises relaxations. \\ \texttt{Will one news cause the other one? \\ 1. No, there is no cause-and-effect relationship between them. \\ 2. The happening of news A will cause news B. \\ 3. The happening of news B will cause news A. \\ Only answer '1' or '2' or '3' without any other words.}} & 1. & 1 & T \\
\midrule
\specialcell{In-Context\\Learning}  & \parbox[c]{9cm}{\texttt{Here are three examples: \\
News A: Why Reliance Industries share price has gained over 19\% in four sessions. News B: IndusInd Bank stock rises over 6\% ahead of Q4 earnings. For this pair of news titles, there is no cause-and-effect relationship between them. \\
News A: Indian government brushes off Indian tax officers' proposal for coronavirus tax on super rich. News B: Inquiry against 50 IRS officers over suggesting tax hike for the rich: Report. For this pair of titles, the happening of news A will cause news B. \\
News A: Insensitive or lost in translation? Twitter weighs in on Thiem's comments against a player fund. News B: Coronavirus: Why should I give money to lower-ranked players, questions Dominic Thiem. For this pair of titles, the happening of news B will cause news A. \\
Now, answer this question. \\
News title A:} Guv encourages creative developers during lockdown. \texttt{News title B:} Govt hints at lockdown extension, but promises relaxations. \texttt{Will one news cause the other one? \\ 1. No, there is no cause-and-effect relationship between them. \\ 2. The happening of news A will cause news B. \\ 3. The happening of news B will cause news A. \\ Only answer '1' or '2' or '3' without any other words.
}} & 2. & 1 & F \\
\bottomrule
\end{tabular}
\caption{Prompt templates used for the HeadlineCause benchmark.}
\label{tab:headlinecause_pemplate}
\end{table*}

\begin{table*}
\small
\centering
\begin{tabular}{c|c|c|c|c}
\toprule
\multicolumn{5}{c}{\textbf{Explicit Discourse Relation Tasks}}                                                   \\
\midrule
\multicolumn{1}{c|}{\textbf{Strategies}} & \textbf{Template input} & \textbf{ChatGPT} & \textbf{Gold} & \textbf{T/F} \\
\midrule
\specialcell{Top-level\\Prompt} & \parbox[c]{8.5cm}{\texttt{Argument 1:}"When used as background in this way, the music has an appropriate eeriness"\\\texttt{Argument 2:}"Served up as a solo the music lacks the resonance provided by a context within another medium"\\\texttt{Connective between Argument 1 and Argument 2:}"however"\\\texttt{Question:What is the discourse relation between Argument 1 and Argument 2?}\\\texttt{A. Comparison}\\\texttt{B. Contingency}\\\texttt{C. Expansion}\\\texttt{D. Temporal}\\\texttt{Answer:}}     &  \scriptsize{B. Contingency}                   & \scriptsize{A. Comparison}         & F            \\
\midrule
\specialcell{Second-level\\Prompt} & \parbox[c]{8.5cm}{\texttt{Argument 1:}"When used as background in this way, the music has an appropriate eeriness"\\\texttt{Argument 2:}"Served up as a solo the music lacks the resonance provided by a context within another medium"\\\texttt{Connective between Argument 1 and Argument 2:}"however"\\\texttt{Question:What is the discourse relation between Argument 1 and Argument 2?}\\\texttt{A. Concession}\\\texttt{B. Contrast}\\\texttt{C. Cause}\\\texttt{D. Condition}\\\texttt{E. Alternative}\\\texttt{F. Conjunction}\\\texttt{G. Instantiation}\\\texttt{H. List}\\\texttt{I. Restatement}\\\texttt{J. Asynchronous}\\\texttt{K. Synchrony}\\\texttt{Answer:}}     & \scriptsize{B. Contrast}                  & \scriptsize{B. Contrast}           & T            \\
\midrule

\specialcell{Prompt\\Engineering} & \parbox[c]{8.5cm}{\texttt{Argument 1:}"When used as background in this way, the music has an appropriate eeriness"\\\texttt{Argument 2:}"Served up as a solo the music lacks the resonance provided by a context within another medium"\\\texttt{Connective between Argument 1 and Argument 2:}"however"\\\texttt{Question:What is the discourse relation between Argument 1 and Argument 2?}\\\texttt{A. Comparison.Concession, nonetheless}\\\texttt{B. Comparison.Contrast, however}\\\texttt{C. Contingency.Cause, so}\\\texttt{D. Contingency.Condition, if}\\\texttt{E. Expansion.Alternative, instead}\\\texttt{F. Expansion.Conjunction, also}\\\texttt{G. Expansion.Instantiation, for example}\\\texttt{H. Expansion.List, and}\\\texttt{I. Expansion.Restatement, specifically}\\\texttt{J. Temporal.Asynchronous, before}\\\texttt{K. Temporal.Synchrony, when}\\\texttt{Answer:}}     &  \scriptsize{\specialcell{B.Comparison.\\Contrast,\\however}} & \scriptsize{\specialcell{B. Comparison.\\Contrast}}        & T            \\

\bottomrule
\end{tabular}
\caption{Prompt example for PDTB2.0 explicit discourse relation task.}
\label{tab:explicitdiscourse_templete_1}
\end{table*}

\begin{table*}
\small
\centering
\begin{tabular}{c|c|c|c|c}
\toprule
\multicolumn{5}{c}{\textbf{Explicit Discourse Relation Tasks}}                                                   \\
\midrule
\multicolumn{1}{c|}{\textbf{Strategies}} & \textbf{Template input} & \textbf{ChatGPT} & \textbf{Gold} & \textbf{T/F} \\
\midrule
\specialcell{In-Context\\Learning} & 
\parbox[c]{8.5cm}{\texttt{All answer select from following:\\A. Comparison.Concession\\B. Comparison.Contrast\\C. Contingency.Cause\\D. Contingency.Condition\\E. Expansion.Alternative\\F. Expansion.Conjunction\\G. Expansion.Instantiation\\H. Expansion.List\\I. Expansion.Restatement\\J. Temporal.Asynchronous\\K. Temporal.Synchrony\\\\Argument 1:"whose hair is thinning and gray and whose face has a perpetual pallor."\\Argument 2:"The prime minister continues to display an energy, a precision of thought and a willingness to say publicly what most other Asian leaders dare say only privately." \\Connective between Argument 1 and Argument 2:"nonetheless"\\Question:What is the discourse relation between Argument 1 and Argument 2?\\Answer:Comparison.Concession\\\\Argument 1:"they usually give current shareholders the right to buy more stock of their corporation at a large discount if certain events occur."\\Argument 2:"these discount purchase rights may generally be redeemed at a nominal cost by the corporation's directors if they approve of a bidder." \\Connective between Argument 1 and Argument 2:"however"\\Question:What is the discourse relation between Argument 1 and Argument 2?\\Answer:Comparison.Contrast\\
......\\
\\Argument 1:"I find it hard to ignore our environmental problems."\\Argument 2:"I start my commute to work with eyes tearing and head aching from the polluted air."\\Connective between Argument 1 and Argument 2:"when" \\Question:What is the discourse relation between Argument 1 and Argument 2?\\Answer:Temporal.Synchrony}\\\\\texttt{Argument 1:}"When used as background in this way, the music has an appropriate eeriness"\\\texttt{Argument 2:}"Served up as a solo the music lacks the resonance provided by a context within another medium"\\\texttt{Connective between Argument 1 and Argument 2:}"however"\\\texttt{Question:What is the discourse relation between Argument 1 and Argument 2?}\\\texttt{Answer:}}     &  \scriptsize{\specialcell{B.Comparison.\\Contrast}} & \scriptsize{\specialcell{B.Comparison.\\Contrast}}        & T            \\

\bottomrule
\end{tabular}
\caption{Prompt example for PDTB2.0 explicit discourse relation task (Continuous).}
\label{tab:explicitdiscourse_templete_2}
\end{table*}

\begin{table*}
\small
\centering
\begin{tabular}{c|c|c|c|c}
\toprule
\multicolumn{5}{c}{\textbf{Implicit Discourse Relation Tasks}}                                                   \\
\midrule
\multicolumn{1}{c|}{\textbf{Strategies}} & \textbf{Template input} & \textbf{ChatGPT} & \textbf{Gold} & \textbf{T/F} \\
\midrule
\specialcell{Top-level\\Prompt} & \parbox[c]{8.5cm}{\texttt{Argument 1:}"We've been spending a lot of time in Los Angeles talking to TV production people"\\\texttt{Argument 2:}"With the competitiveness of the television market these days, everyone is looking for a way to get viewers more excited"\\\texttt{Question:What is the discourse relation between Argument 1 and Argument 2?}\\\texttt{A. Comparison}\\\texttt{B. Contingency}\\\texttt{C. Expansion}\\\texttt{D. Temporal}\\\texttt{Answer:}} & \scriptsize{\specialcell{C. Expansion}}  & \scriptsize{\specialcell{B. Contingency}}  & F\\

\midrule
\specialcell{Second-level\\Prompt} & \parbox[c]{8.5cm}{\texttt{Argument 1:}"We've been spending a lot of time in Los Angeles talking to TV production people"\\\texttt{Argument 2:}"With the competitiveness of the television market these days, everyone is looking for a way to get viewers more excited"\\\texttt{Question:What is the discourse relation between Argument 1 and Argument 2?}\\\texttt{A. Concession}\\\texttt{B. Contrast}\\\texttt{C. Cause}\\\texttt{D. Pragmatic Cause}\\\texttt{E. Alternative}\\\texttt{F. Conjunction}\\\texttt{G. Instantiation}\\\texttt{H. List}\\\texttt{I. Restatement}\\\texttt{J. Asynchronous}\\\texttt{K. Synchrony}\\\texttt{Answer:}}    & \scriptsize{\specialcell{C. Cause}}    & \scriptsize{\specialcell{C. Cause}}         & T        \\
\midrule

\specialcell{Prompt\\Engineering} & \parbox[c]{8.5cm}{\texttt{Argument 1:}"We've been spending a lot of time in Los Angeles talking to TV production people"\\\texttt{Argument 2:}"With the competitiveness of the television market these days, everyone is looking for a way to get viewers more excited"\\\texttt{Question:What is the discourse relation between Argument 1 and Argument 2?}\\\texttt{A. Comparison.Concession, if}\\\texttt{B. Comparison.Contrast, however}\\\texttt{C. Contingency.Cause, so}\\\texttt{D. Contingency.Pragmatic, indeed}\\\texttt{E. Expansion.Alternative, instead}\\\texttt{F. Expansion.Conjunction, also}\\\texttt{G. Expansion.Instantiation, for example}\\\texttt{H. Expansion.List, and}\\\texttt{I. Expansion.Restatement, specifically}\\\texttt{J. Temporal.Asynchronous, before}\\\texttt{K. Temporal.Synchrony, when}\\\texttt{Answer:}}   & \scriptsize{\specialcell{C. Contingency.\\Cause,\\so}}                   & \scriptsize{\specialcell{C. Contingency.\\Cause}}        & T            \\

\bottomrule
\end{tabular}
\caption{Prompt example for PDTB2.0 implicit discourse relation task.}
\label{tab:implicitdiscourse_templete_1}
\end{table*}

\begin{table*}
\small
\centering
\begin{tabular}{c|c|c|c|c}
\toprule
\multicolumn{5}{c}{\textbf{Implicit Discourse Relation Tasks}}                                                   \\
\midrule
\multicolumn{1}{c|}{\textbf{Strategies}} & \textbf{Template input} & \textbf{ChatGPT} & \textbf{Gold} & \textbf{T/F} \\
\midrule
\specialcell{In-Context\\Learning} & \parbox[c]{8.5cm}{\texttt{All answer select from following:\\A. Comparison.Concession, nonetheless\\B. Comparison.Contrast, however\\C. Contingency.Cause, so\\D. Contingency.Pragmatic Cause, indeed\\E. Expansion.Alternative, instead\\F. Expansion.Conjunction, also\\G. Expansion.Instantiation, for example\\H. Expansion.List, and\\I. Expansion.Restatement, specifically\\J. Temporal.Asynchronous, before\\K. Temporal.Synchrony, when\\\\Argument 1:"Coke could be interested in more quickly developing some of the untapped potential in those markets."\\Argument 2:"A Coke spokesman said he couldn't say whether that is the direction of the talks." \\Question:What is the discourse relation between Argument 1 and Argument 2?\\Answer:Comparison.Concession, nonetheless\\\\Argument 1:"Tanks currently are defined as armored vehicles weighing 25 tons or more that carry large guns."\\Argument 2:"The Soviets complicated the issue by offering to include light tanks, which are as light as 10 tons." \\Question:What is the discourse relation between Argument 1 and Argument 2?\\Answer:Comparison.Contrast, however\\
......\\
\\Argument 1:"Panamanian dictator Torrijos, he was told, had granted the shah of Iran asylum in Panama as a favor to Washington."\\Argument 2:"Mr.Sanford was told Mr.Noriega's friend, Mr. Wittgreen, would be handling the shah's security." \\Question:What is the discourse relation between Argument 1 and Argument 2?\\Answer: Temporal.Synchrony, when}\\\\\texttt{Argument 1:}"We've been spending a lot of time in Los Angeles talking to TV production people"\\\texttt{Argument 2:}"With the competitiveness of the television market these days, everyone is looking for a way to get viewers more excited"\\\texttt{Question:What is the discourse relation between Argument 1 and Argument 2?}\\\texttt{Answer:}}                       & \scriptsize{\specialcell{C. Contingency.\\Cause,\\so}}                   & \scriptsize{\specialcell{C. Contingency.\\Cause}}        & T      \\
\bottomrule
\end{tabular}
\caption{Prompt example for PDTB2.0 implicit discourse relation task (Continued).}
\label{tab:implicitdiscourse_templete_2}
\end{table*}

\begin{table*}
\small
\centering
\begin{tabular}{l|c|c|c|c}
\toprule
\multicolumn{5}{c}{\textbf{DiscoGeM}}                                                   \\
\midrule
\multicolumn{1}{c|}{\textbf{Strategies}} & \textbf{Template input} & \textbf{ChatGPT} & \textbf{Gold} & \textbf{T/F} \\
\midrule
\specialcell{Prompt} & \parbox[c]{8.5cm}{\texttt{Argument 1:}"Allow me to make a few general comments on European solidarity, on the Solidarity Fund and on some events that may provide lessons for the future."\\\texttt{Argument 2:}"In 2002 I had the experience of leading a country that was struck by terrible floods, together with the Federal Republic of Germany and Austria. It was the scale of that disaster that provided the incentive for the creation of the Solidarity Fund."\\\texttt{Question:What is the discourse relation between Argument 1 and Argument 2?}\\\texttt{(0) arg1-as-denier}\\\texttt{(1) arg1-as-detail}\\\texttt{(2) arg1-as-goal}\\\texttt{(3) arg2-as-denier}\\\texttt{(4) arg2-as-detail}\\\texttt{(5) arg2-as-goal}\\\texttt{(6) arg2-as-instance}\\\texttt{(7) arg2-as-subst}\\\texttt{(8) conjunction}\\\texttt{(9) contrast}\\\texttt{(10) differentcon}\\\texttt{(11) disjunction}\\\texttt{(12) precedence}\\\texttt{(13) reason }\\\texttt{(14) result}\\\texttt{(15) similarity}\\\texttt{(16) succession}\\\texttt{(17) synchronous}\\\texttt{Answer:}} & \scriptsize{\specialcell{(2) arg1-as-goal}}  & \scriptsize{\specialcell{(4) arg2-as-detail}}  & F\\
\bottomrule
\end{tabular}
\caption{Prompt example 1 for DiscoGeM, the multi-genre discourse classification task.}
\label{tab:discogem-example-1}
\end{table*}

\begin{table*}
\small
\centering
\begin{tabular}{l|c|c|c|c}
\toprule
\multicolumn{5}{c}{\textbf{DiscoGeM}}                                                   \\
\midrule
\multicolumn{1}{c|}{\textbf{Strategies}} & \textbf{Template input} & \textbf{ChatGPT} & \textbf{Gold} & \textbf{T/F} \\

\midrule
\specialcell{Prompt\\Engineering} & \parbox[c]{8.5cm}{\texttt{Argument 1:}"However, the Member States are not obliged to replace fixed-term contracts with open-ended contracts assuming that there are other effective measures in place that would prevent or sanction such abuse. The European Court of Justice confirmed this interpretation in its judgment of 4 July 2006 in Case C-212/04 (Adeneler) pertaining to Greek legislation."\\\texttt{Argument 2:}"The European Court of Justice also stated that interpretation of the relevant national legislation does not fall within its competence. It is entirely for the Greek courts to provide an interpretation of relevant Greek legislation and to determine whether this legislation complies with the requirements of the Directive regarding the existence of effective measures that would prevent and sanction abuse arising from the use of successive fixed-term employment contracts."\\\texttt{Question:What is the discourse relation between Argument 1 and Argument 2?}\\\texttt{(0) arg1-as-denier: despite the fact that}\\\texttt{(1) argument 1 as detail: in short}\\\texttt{(2) argument 1 as goal: for that purpose}\\\texttt{(3) argument 2 as denier: despite this}\\\texttt{(4) argument 2 as detail: in more detail}\\\texttt{(5) argument 2 as goal: ensuring that}\\\texttt{(6) argument 2 as instance: for instance}\\\texttt{(7) argument 2 as substitution: rather}\\\texttt{(8) conjunction: in addition}\\\texttt{(9) contrast: by comparison}\\\texttt{(10) differentcon: none}\\\texttt{(11) disjunction: or alternatively}\\\texttt{(12) precedence: subsequently}\\\texttt{(13) reason: the reasons is/are that }\\\texttt{(14) result: consequently}\\\texttt{(15) similarity: similarly}\\\texttt{(16) succession: previously}\\\texttt{(17) synchronous: at that time}\\\texttt{Answer:}}    & \scriptsize{\specialcell{(8) conjunction: \\ in addition}}    & \scriptsize{\specialcell{(8) conjunction: \\ in addition}}      & T        \\

\bottomrule
\end{tabular}
\caption{Prompt example 2 for DiscoGeM, the multi-genre discourse classification task.}
\label{tab:discogem-example-2}
\end{table*}

\begin{table*}
\small
\centering
\begin{tabular}{l|c|c|c|c}
\toprule
\multicolumn{5}{c}{\textbf{DiscoGeM}}                                                   \\
\midrule
\multicolumn{1}{c|}{\textbf{Strategies}} & \textbf{Template input} & \textbf{ChatGPT} & \textbf{Gold} & \textbf{T/F} \\

\midrule
\specialcell{In-Context\\Learning} & \parbox[c]{8.5cm}{\texttt{Candidate relations:}\\\texttt{(0) arg1-as-denier: despite the fact that}\\\texttt{(1) argument 1 as detail: in short}\\\texttt{(2) argument 1 as goal: for that purpose}\\\texttt{(3) argument 2 as denier: despite this}\\\texttt{(4) argument 2 as detail: in more detail}\\\texttt{(5) argument 2 as goal: ensuring that}\\\texttt{(6) argument 2 as instance: for instance}\\\texttt{(7) argument 2 as substitution: rather}\\\texttt{(8) conjunction: in addition}\\\texttt{(9) contrast: by comparison}\\\texttt{(10) differentcon: none}\\\texttt{(11) disjunction: or alternatively}\\\texttt{(12) precedence: subsequently}\\\texttt{(13) reason: the reasons is/are that }\\\texttt{(14) result: consequently}\\\texttt{(15) similarity: similarly}\\\texttt{(16) succession: previously}\\\texttt{(17) synchronous: at that time}\\\texttt{Argument 1:}"Mr President, ladies and gentlemen, the motion for a resolution before us today is important because of its subject and the desire to protect the rule of law and press freedom. It is also very important because of the broad consensus which has finally been reached after some heated discussions behind the scenes."\\\texttt{Argument 2:}"The problem considered by the motion is a major one but, as has already been touched upon, could be regarded as minor in light of the even greater problem of the general situation in Angola which is experiencing a terrible humanitarian disaster. This situation, as in neighbouring former Zaire, is like a festering wound in which it is not clear who is infecting whom."\\\texttt{Question:What is the discourse relation between Argument 1 and Argument 2?}\\\texttt{Answer: (14) result: consequently}\\\texttt{Argument 1:}"The ship was finally able to turn around and it fled northwards as fast as possible. Then there was a terrible explosion about six hundred yards to the stern and a gigantic column of water and steam, perhaps a hundred yards high, shot out of the sea. The Oudenbourg set course for Harwich and sent out a radio warning in all directions: \"Attention all shipping, attention all shipping!"\\\texttt{Argument 2:}"Severe danger on Ostende-Ramsgate lane. Underwater explosion. Cause unknown. All shipping advised avoid area!"\\\texttt{Question:What is the discourse relation between Argument 1 and Argument 2?}\\\texttt{Answer: (4) argument 2 as detail: in more detail}\\\texttt{Argument 1:}"Allow me to make a few general comments on European solidarity, on the Solidarity Fund and on some events that may provide lessons for the future. In 2002 I had the experience of leading a country that was struck by terrible floods, together with the Federal Republic of Germany and Austria."\\\texttt{Argument 2:}"It was the scale of that disaster that provided the incentive for the creation of the Solidarity Fund. The disaster occurred in August and the first payments were received by the Czech Republic the following January."\\\texttt{Question:What is the discourse relation between Argument 1 and Argument 2?}\\\texttt{Answer:}}    & \scriptsize{\specialcell{(3) argument 2 \\as denier:\\despite this}}    & \scriptsize{\specialcell{(14) result:\\consequently}}      & F       \\

\bottomrule
\end{tabular}
\caption{Prompt example 3 for DiscoGeM, the multi-genre discourse classification task.}
\label{tab:discogem-example-3}
\end{table*}

\begin{table*}
\small
\centering
\begin{tabular}{l|c|c|c}
\toprule
\multicolumn{4}{c}{\textbf{Dialogue DP-STAC}}                                                   \\
\midrule
\multicolumn{1}{c|}{\textbf{Strategies}} & \textbf{Template input} & \textbf{ChatGPT} & \textbf{Gold} \\

\midrule
\specialcell{Prompt\\(w/o desc.)} & \parbox[c]{8.5cm}{\texttt{Here is a multi-party dialogue:}\\Utterance 0: (Speaker A) sorry raef-\\Utterance 1: (Speaker A) at least i forgot to play it\\Utterance 2: (Speaker A) before that 6 was rolled\\Utterance 3: (Speaker B) well at least people should realize your advantage now\\\\\texttt{Q: Predict all the possible discourse relations between utterances and their types line by line\\(e.g., 'Utterance 0 and utternace 1: (0)\\Utterance 0 and utterance 3: (1)')\\Candidate types are listed below:\\Choose from:\\(0) Comment\\(1) Clarification question\\(2) Question-answer pair\\(3) Continuation\\(4) Acknowledgement\\(5) Question and elaboration\\(6) Result\\(7) Elaboration\\(8) Explanation\\(9) Correction\\(10) Contrast\\(11) Conditional\\(12) Background\\(13) Narration\\(14) Alternation\\(15) Parallel}}
& \scriptsize{\specialcell{Utterance 0 and\\utterance 1: (2)\\\quad\\Utterance 1 and\\utterance 2: (0)\\\quad\\Utterance 2 and\\utterance 3: (9)\\\quad\\Utterance 3 and\\utterance 4: (0)\\\quad\\Utterance 4 and\\utterance 5: (5)\\\quad\\Utterance 5 and\\utterance 6: (0)\\\quad\\Utterance 6 and\\utterance 7: (7)\\\quad\\Utterance 7 and\\utterance 8: (0)\\\quad\\Utterance 8 and\\utterance 9: (3)\\\quad\\Utterance 9 and\\utterance 10: (14)}}    & \scriptsize{\specialcell{Utterance 0 and\\utterance 1: (8)\\\quad\\Utterance 1 and\\utterance 2: (13)\\\quad\\Utterance 1 and\\utterance 3: (0)}}              \\\\
\bottomrule
\end{tabular}
\caption{Prompt example for STAC in the multi-party dialogue discourse parsing task. Examples in Molweni are in a similar format.}
\label{tab:stac-example-1}
\end{table*}

\begin{table*}
\small
\centering
\begin{tabular}{l|c|c|c}
\toprule
\multicolumn{4}{c}{\textbf{Dialogue DP-STAC}}                                                   \\
\midrule
\multicolumn{1}{c|}{\textbf{Strategies}} & \textbf{Template input} & \textbf{ChatGPT} & \textbf{Gold} \\
\midrule
\specialcell{Prompt\\(w/ desc.)} & \parbox[c]{8cm}{\texttt{Here is a multi-party dialogue:}\\Utterance 0: (Speaker A) sorry raef-\\Utterance 1: (Speaker A) at least i forgot to play it\\Utterance 2: (Speaker A) before that 6 was rolled\\Utterance 3: (Speaker B) well at least people should realize your advantage now\\\\\texttt{Q: Predict all the possible discourse relations between utterances and their types line by line\\(e.g., 'Utterance 0 and utternace 1: (0)\\Utterance 0 and utterance 3: (1)')\\Candidate types are listed below:\\Choose from:\\(0) Comment: Utterance y comments utterance x.\\(1) Clarification question: Utterance y clarifies utterance x.\\(2) Question-answer pair: Utterance x is a question and utterance y is the answer of utterance x.\\(3) Continuation: Utterance y is the continuation of utterance x.\\(4) Acknowledgement: Utterance y acknowledges utterance x.\\(5) Question and elaboration: Utterance x is a question and utterance y tries to elaborate utterance x.\\(6) Result: Utterance y is the effect brought about by the situation described in utterance x.\\(7) Elaboration: Utterance y elaborates utterance x.\\(8) Explanation: Utterance y is the explanation of utterance x.\\(9) Correction: Utterance y corrects utterance x.\\(10) Contrast: Utterance x and utterance y share a predicate or property and a difference on shared property.\\(11) Conditional: Utterance x is the condition of utterance y or utterance y is the condition of utterance x.\\(12) Background: Utterance y is the background of utterance x.\\(13) Narration: Utterance y is the narration of utterance x.\\(14) Alternation: Utterance x and utterance y denote alternative situations.\\(15) Parallel: Utterance y and utterance x are parallel and present almost the same meaning.}}
& \scriptsize{\specialcell{Utterance 0 and\\utterance 1: (2)\\\quad\\Utterance 0 and\\utterance 3: (1)\\\quad\\Utterance 1 and\\utterance 5: (0)\\\quad\\Utterance 2 and\\utterance 3: (4)\\\quad\\Utterance 4 and\\utterance 5: (0)\\\quad\\Utterance 6 and\\utterance 7: (4)\\\quad\\Utterance 8 and\\utterance 9: (0)\\\quad\\Utterance 9 and\\utterance 10: (9)}}    & \scriptsize{\specialcell{Utterance 0 and\\utterance 1: (8)\\\quad\\Utterance 1 and\\utterance 2: (13)\\\quad\\Utterance 1 and\\utterance 3: (0)}}              \\\\
\bottomrule
\end{tabular}
\caption{Prompt example for STAC in the multi-party dialogue discourse parsing task. Examples in Molweni are in the similar format.}
\label{tab:stac-example-2}
\end{table*}

\begin{table*}
\small
\centering
\begin{tabular}{l|c|c|c}
\toprule
\multicolumn{4}{c}{\textbf{Dialogue DP-STAC}}                                                   \\
\midrule
\multicolumn{1}{c|}{\textbf{Strategies}} & \textbf{Template input} & \textbf{ChatGPT} & \textbf{Gold} \\

\midrule
\specialcell{In-Context\\Learning} & \parbox[c]{8cm}{\texttt{[Example 1]}\\\texttt{Here is a multi-party dialogue:}\\Utterance 0: (Speaker A) sorry raef-\\Utterance 1: (Speaker A) at least i forgot to play it\\Utterance 2: (Speaker A) before that 6 was rolled\\Utterance 3: (Speaker B) well at least people should realize your advantage now\\\\\texttt{Q: Predict all the possible discourse relations between utterances and their types line by line\\(e.g., 'Utterance 0 and utternace 1: (0)\\Utterance 0 and utterance 3: (1)')\\Candidate types are listed below:\\Choose from:\\(0) Comment\\(1) Clarification question\\(2) Question-answer pair\\(3) Continuation\\(4) Acknowledgement\\(5) Question and elaboration\\(6) Result\\(7) Elaboration\\(8) Explanation\\(9) Correction\\(10) Contrast\\(11) Conditional\\(12) Background\\(13) Narration\\(14) Alternation\\(15) Parallel\\\\A:}\\Utterance 0 and utterance 1: (8)\\Utterance 1 and utterance 2: (13)\\Utterance 1 and utterance 3: (0)\\\\\texttt{[Example 2]}\\\texttt{Here is a multi-party dialogue:}\\Utterance 0: (Speaker A) I need wood, clay or ore,  I can give Sheep\\Utterance 1: (Speaker B) i can trade wood\\Utterance 2: (Speaker C) just spent it all\\Utterance 3: (Speaker C) sorry\\Utterance 4: (Speaker A) 1 sheep for 1 wood?\\Utterance 5: (Speaker B) 2 sheep 1 wood\\Utterance 6: (Speaker C) sorry empty\\Utterance 7: (Speaker C) tough times..\\Utterance 8: (Speaker B) hopefully i dont roll a 7\\Utterance 9: (Speaker B) and that biotes me in the arse\\Utterance 10: (Speaker B) bites*\\\\\texttt{Q: Predict all the possible discourse relations between utterances and their types line by line\\(e.g., 'Utterance 0 and utternace 1: (0)\\Utterance 0 and utterance 3: (1)')\\Candidate types are listed below:\\Choose from:\\(0) Comment\\(1) Clarification question\\(2) Question-answer pair\\(3) Continuation\\{\color{blue} (... same as above)}\\(14) Alternation\\(15) Parallel\\}}
& \scriptsize{\specialcell{A:\\Utterance 0 and\\utterance 1: (2)\\\quad\\Utterance 1 and\\utterance 4: (2)\\\quad\\Utterance 2 and\\utterance 3: (9)\\\quad\\Utterance 4 and\\utterance 5: (5)\\\quad\\Utterance 5 and\\utterance 6: (7)\\\quad\\Utterance 5 and\\utterance 8: (3)\\\quad\\Utterance 8 and\\utterance 9: (11)\\\quad\\Utterance 9 and\\utterance 10: (9)}}    & \scriptsize{\specialcell{A:\\Utterance 0 and\\utterance 1: (2)\\\quad\\Utterance 0 and\\utterance 2: (2)\\\quad\\Utterance 2 and\\utterance 3: (0)\\\quad\\Utterance 1 and\\utterance 4: (5)\\\quad\\Utterance 4 and\\utterance 5: (2)\\\quad\\Utterance 6 and\\utterance 7: (8)\\\quad\\Utterance 8 and\\utterance 9: (3)\\\quad\\Utterance 9 and\\utterance 10: (9)\\\quad\\Utterance 2 and\\utterance 6: (7)\\\quad\\Utterance 5 and\\utterance 8: (0)}}              \\\\
\bottomrule
\end{tabular}
\caption{Prompt example for STAC in the multi-party dialogue discourse parsing task. Examples in Molweni are in the similar format.}
\label{tab:stac-example-3}
\end{table*}

\begin{table*}
\small
\centering
\begin{tabular}{c|c|c|c|c}
\toprule
\multicolumn{5}{c}{\textbf{CKBP}}                                                   \\
\midrule
\multicolumn{1}{c|}{\textbf{Strategies}} & \textbf{Template input} & \textbf{ChatGPT} & \textbf{Gold} & \textbf{T/F} \\
\midrule
\specialcell{Prompt\\Engineering}  & \parbox[c]{9cm}{\texttt{Answer whether the following statement is plausible. Answer with only Yes or No:If} PersonX drinks coffee, \texttt{as a result, PersonX feels,} refreshed.} & Yes & Yes & T \\
\midrule
\specialcell{In-Context\\Learning}  & \parbox[c]{9cm}{\texttt{\texttt{Answer whether the following statement is plausible. Answer with only Yes or No:If} PersonY accept the interview, as a result, PersonY or others will, PersonX give PersonY this opportunity.A: No\\\texttt{Answer whether the following statement is plausible. Answer with only Yes or No:If} PersonX lead the line, as a result, PersonY or others feel, PersonX support PersonX family.A: No\\\texttt{Answer whether the following statement is plausible. Answer with only Yes or No:If} PersonX form PersonY conception, as a result, PersonY or others want to, PersonY want to discuss with PersonZ.A: Yes\\\texttt{Answer whether the following statement is plausible. Answer with only Yes or No:If} PersonX give, PersonX is seen as, PersonX be comunicative.A: Yes\\\texttt{Answer whether the following statement is plausible. Answer with only Yes or No:If} PersonX be nervous, as a result, PersonX will, that be important to PeopleX.A: Yes\\\texttt{Answer whether the following statement is plausible. Answer with only Yes or No:If} PersonX celebrate persony, because PersonX wanted, PersonX feel oneself.A: No\\\texttt{Answer whether the following statement is plausible. Answer with only Yes or No:If} PersonX learn to ride a bike, but before, PersonX needed, PersonX wear helmet.A: Yes\\\texttt{Answer whether the following statement is plausible. Answer with only Yes or No:If} PersonX take PersonY time, as a result, PersonX feels, PersonX feel mortified.A: No\\\texttt{Answer whether the following statement is plausible. Answer with only Yes or No:If} PersonX want to ask a tought question, as a result, PersonX wants to, PersonX want to throw out PersonX clothes.A: No\\\texttt{Answer whether the following statement is plausible. Answer with only Yes or No:If} PersonX achieve PersonX end, happens after, PersonX start a small business.A: Yes\\\texttt{Answer whether the following statement is plausible. Answer with only Yes or No:If} PersonX like the idea, happens before, PersonX call a uber.A: No\\\texttt{Answer whether the following statement is plausible. Answer with only Yes or No:If} PersonX get injure, because, PersonX feel odd.A: No\\\texttt{Answer whether the following statement is plausible. Answer with only Yes or No:If} person x be bed ridden with illness, can be hindered by, PersonX find the perfect dog.A: No\\\texttt{Answer whether the following statement is plausible. Answer with only Yes or No:If} PersonX play violin, includes the event or action, PersonX make noise.A: Yes\\\texttt{Answer whether the following statement is plausible. Answer with only Yes or No:If} PersonX could not complete something, causes, PeopleX have find it.A: No\\\texttt{Answer whether the following statement is plausible. Answer with only Yes or No:If}} PersonX drinks coffee, \texttt{as a result, PersonX feels,} refreshed.} & Yes & Yes & T \\
\bottomrule
\end{tabular}
\caption{Prompt example for CKBP.}
\label{tab:ckbp_pemplate}
\end{table*}

\begin{table*}
\small
\centering
\begin{tabular}{c|c|c|c|c}
\toprule
\multicolumn{5}{c}{\textbf{DiscoSense}}                                                   \\
\midrule
\multicolumn{1}{c|}{\textbf{Strategies}} & \textbf{Template input} & \textbf{ChatGPT} & \textbf{Gold} & \textbf{T/F} \\
\midrule
\specialcell{Prompt\\Engineering}  & \parbox[c]{8.5cm}{\texttt{Question:} Which option represents the most plausible ending of the given context?\\
\texttt{Context:} Although it took a while to assemble, the instructions are easy to follow. \textit{overall}\\
\texttt{Option 1:} This tv stand is worth purchasing for.\\
\texttt{Option 2:} The dining room set is a quality item that will last for the only thing I will complain about was the fact that there was dust in the boxes.\\
\texttt{Option 3:} The stool works well for our needs.\\
\texttt{Option 4:} The desk took less than 1 hour to assemble and has a contemporary look with espresso-colored legs.\\
\texttt{Select only from ["Option 1", "Option 2", "Option 3", "Option 4"]}} & Option 4 & Option 1 & F\\
\midrule
\specialcell{In-Context\\Learning} & \parbox[c]{8.5cm}{\texttt{Question: Which option represents the most plausible ending of the given context?\\
Context: Both sides have in their own way proved themselves as bad as each other. in short\\
\texttt{Option 1:} The problem is not the attitude of individual men but the spirit of the times.\\
\texttt{Option 2:} The us government has been taken over by and both by corporate interests and political hacks.\\
\texttt{Option 3:} You have a society that has been utterly corrupted by money and power.\\
\texttt{Option 4:} Blacklisting worked against labour in wales, in london, and possibly, if he tries it in scotland, it will rebound there.\\
\texttt{Select only from ["Option 1", "Option 2", "Option 3", "Option 4"]}\\
Option 4\\\\
\texttt{Question:} Which option represents the most plausible ending of the given context?\\
Context: Any trinidadian wanting to vote must prove they maintain a residence there. because of that\\
\texttt{Option 1:} No one living in the streets of burlington will ever be allowed to vote.\\
\texttt{Option 2:} They are not eligible to vote.\\
\texttt{Option 3:} And because the official election is open to all, the town hall will remain open for voting on election day.\\
\texttt{Option 4:} Most trinidadians living here wont be able to vote.\\
\texttt{Select only from ["Option 1", "Option 2", "Option 3", "Option 4"]}\\
Option 4\\\\
...}\\\\
\texttt{Question:} Which option represents the most plausible ending of the given context?\\
\texttt{Context:} Although it took a while to assemble, the instructions are easy to follow. \textit{overall}\\
\texttt{Option 1:} This tv stand is worth purchasing for.\\
\texttt{Option 2:} The dining room set is a quality item that will last for the only thing I will complain about was the fact that there was dust in the boxes.\\
\texttt{Option 3:} The stool works well for our needs.\\
\texttt{Option 4:} The desk took less than 1 hour to assemble and has a contemporary look with espresso-colored legs.\\
\texttt{Select only from ["Option 1", "Option 2", "Option 3", "Option 4"]}} & Option 1 & Option 1 & T\\
\bottomrule
\end{tabular}
\caption{Prompt example for DiscoSense.}
\label{tab:discosense_template}
\end{table*}


\end{document}